\pdfoutput=1

\documentclass[11pt]{article}

\usepackage[preprint]{acl}

\usepackage{times}
\usepackage{latexsym}

\usepackage[T1]{fontenc}

\usepackage[utf8]{inputenc}

\usepackage{microtype}

\usepackage{inconsolata}

\usepackage{graphicx}

\usepackage{amsmath}
\usepackage{bbding}
\usepackage{booktabs} 
\usepackage{multirow}
\usepackage{multicol}

\usepackage{array}

\usepackage{tcolorbox}

\definecolor{red}{RGB}{253,93,93}
\definecolor{green}{RGB}{5,146,18} 
\definecolor{venn_red}{RGB}{232,86,66}
\definecolor{venn_orange}{RGB}{240,135,5}

\title{\textit{Confidence v.s. Critique:} \\ A Decomposition of Self-Correction Capability for LLMs}

\author{Zhe Yang$^{1}$, Yichang Zhang$^{2}$, Yudong Wang$^{1}$, Ziyao Xu$^{1}$, Junyang Lin$^{2}$, Zhifang Sui$^{1}$ \\
$^1$State Key Laboratory of Multimedia Information Processing, \\
School of Computer Science, Peking University \\ 
$^2$Alibaba Group \\
\texttt{\{yz\_young, szf\}@pku.edu.cn}\\
}

\begin{document}
\maketitle
\begin{abstract}

Large Language Models (LLMs) can correct their self-generated responses, but a decline in accuracy after self-correction is also witnessed. To have a deeper understanding of self-correction,  we endeavor to decompose, evaluate, and analyze the self-correction behaviors of LLMs.
By enumerating and analyzing answer correctness before and after self-correction, we decompose the self-correction capability into confidence (being confident to correct answers) and critique (turning wrong answers to correct) capabilities, and propose two metrics from a probabilistic perspective to measure these 2 capabilities, along with another metric for overall self-correction capability evaluation.
Based on our decomposition and evaluation metrics, we conduct extensive experiments and draw some empirical conclusions. 
For example, we find different models can exhibit distinct behaviors: some models are confident while others are more critical.
We also find the trade-off between the two capabilities (i.e. improving one can lead to a decline in the other) when manipulating model self-correction behavior by prompts or in-context learning. 
Further, we find a simple yet efficient strategy to improve self-correction capability by transforming Supervision Fine-Tuning (SFT) data format, and our strategy outperforms vanilla SFT in both capabilities and achieves much higher accuracy after self-correction.
Our code will be publicly available on GitHub.
\footnote{\url{https://github.com/Zhe-Young/SelfCorrectDecompose}}

\end{abstract}

\section{Introduction}
\label{sec:intrudcution}
With the increase of training corpus and the number of parameters \citep{radford2018improving,radford2019language,brown2020language}, LLMs have shown remarkable performance in various tasks, but it remains challenging to avoid generating incorrect answers. 
One approach for better performance is \textit{intrinsic self-correction} \citep{kamoi-etal-2024-llms,pan-etal-2024-automatically}, which allows the model to check and revise its self-generated answers without external feedback \citep{wu-etal-2024-large,xi-etal-2023-self}, and this process is quite analogous to human thinking.
\citet{madaan2024self,liu2024large} find self-correction can lead to better responses at the cost of increased inference time \citep{qurecursive}, significantly enhancing model performance. 
However, negative opinions on self-correction also exist \citep{huanglarge,jiang2024self,valmeekam2023can}, and \citet{stechly2023gpt,tyen-etal-2024-llms,jiang2024self} find LLMs even can not determine the correctness of answers, as they often turn correct answers to incorrect ones or fail to correct erroneous answers.
The debate in previous work indicates a lack of deeper understanding of self-correction. 
To narrow this gap, we propose a methodology to decompose, evaluate, analyze, and improve the self-correction capability of LLMs.

\textbf{Self-correction decomposition.} In \S \ref{sec:decomposition}, we enumerate the correctness of answers before and after self-correction and analyze four scenarios, 
based on which we decompose the self-correction capability into: 1. confidence capability (maintaining confidence in correct answers) and 2. critique capability (turning wrong answers to correct).

\begin{figure*}[!t]
    \centering
    \includegraphics[width=0.95\textwidth]{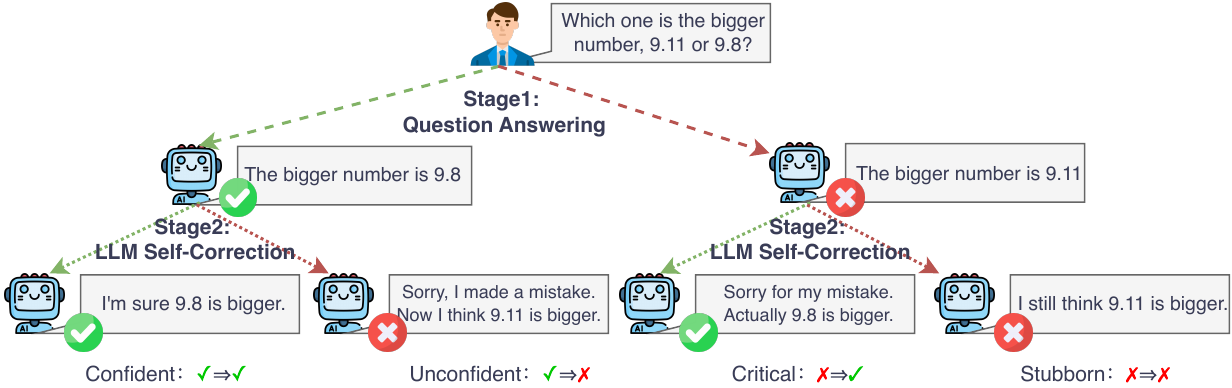}
    \caption{An example of four scenarios in self-correction. For a correct initial answer, LLM can (1). confidently maintain it or (2). unconfidently change it into a wrong answer. For a wrong initial answer, LLM can (3). critique and make it correct or (4). stubbornly insist the wrong answer.}
    \label{fig:4class}
\end{figure*}

\textbf{Self-correction evaluation.} To measure these two capabilities, in \S \ref{sec:methodology} we introduce Confidence Level (CL) and Critique Score (CS) from a probabilistic perspective, which respectively represent the conditional probabilities of the model generating a correct answer after self-correction, given the initial answer is correct/incorrect. 
We also mathematically prove that the accuracy after self-correction can essentially be seen as a weighted sum of these two metrics, which further validates the rationality of our decomposition. 
By analyzing lower and upper bounds of CL and CS, we propose Relative Self-correction Score to measure the overall self-correction capability. 
The calculation of proposed metrics relies on event probabilities, so we further provide probability estimation methods for both classification and generation tasks.

\textbf{Self-correction analysis.} Based on our proposed metrics, in \S \ref{sec:experiments} we conduct extensive experiments across a variety of models
and find that: 1. self-correction usually but not necessarily leads to higher performance; 2. confidence capability is generally better than critique capability for most models; 3. different models can exhibit distinct behaviors; some models are "conservative" (high CL and low CS) while others are more "liberal" (low CL and high CS); 4. models from the same series tend to behave similarly, which may because of their similar pre-training corpus.
In \S \ref{sec:Analysis}, we attempt to manipulate self-correction behaviors of LLMs by prompting \citep{li2024confidence,huanglarge} and in-context learning (ICL) \citep{dong-etal-2024-survey},
finding that simultaneous enhancement in both capabilities can hardly be achieved without fine-tuning, and improving one capability often leads to a decline in the other. 

\textbf{Self-correction improvement.} Based on the above findings and analysis, in \S \ref{sec:tuning} we propose Confidence and Critique improvement Tuning (CCT), a simple yet efficient training strategy to improve self-correction capability of LLMs.
Unlike vanilla SFT, which directly teaches the model a correct answer with the question as context, CCT utilizes the question along with initial correct/incorrect answers as context and teaches model the final answer, enabling the model to maintain correct answers and refine wrong answers.
Experimental results demonstrate that CCT outperforms SFT by a large margin on accuracy after self-correction, breaking the trade-off and achieving higher both CL and CS.

Our contributions can be summarized as follows:
\begin{enumerate}
\item We decompose self-correction capability into confidence and critique capacities, and introduce two metrics to measure them, along with another metric to measure overall self-correction capability.
\item Based on our proposed metrics and probability estimation methods, we conduct extensive experiments across a variety of LLMs and draw some empirical conclusions.
\item We also find confidence and critique capacities can hardly be improved simultaneously through prompting or ICL, and further analyze the trade-off between them.
\item We propose CCT, a simple yet efficient training method to improve self-correction capability, outperforming SFT in both aspects.
\end{enumerate}

\section{Self-Correction Decomposition}

\label{sec:decomposition}
According to different settings discussed in \citet{kamoi-etal-2024-llms}, the self-correction we study can be categorized as \textit{post-hoc intrinsic self-correction}, where LLMs can review and refine their generated responses without external feedback and then output the revised final answers.
Since there is no standard verifier to determine the correctness of a generated answer during this process, the model should first determine whether the answer is correct by itself. If deemed correct, the model persists in outputting it; if considered incorrect, the model then adjusts and outputs a revised answer. 
We divide the process before and after self-correction into two phases:\\
\textbullet~Phase 1 (Question Answering): a question is fed into the model and an answer that can be either correct or incorrect is generated. \\
\textbullet~Phase 2 (Self-Correction): the model is instructed to correct its answer and output a revised answer that also can be correct or incorrect. 

Similar to \citet{zhang-etal-2024-self-contrast}, by considering the Cartesian product of the outcomes from these two phases we categorize four scenarios (as illustrated in Figure \ref{fig:4class}):
\begin{enumerate}
\item \textbf{Confident} (\Checkmark $\rightarrow$ \Checkmark): The model initially generates a correct answer and confidently maintains this correct answer.
\item \textbf{Unconfident} (\Checkmark $\rightarrow$ \XSolidBrush): The model initially generates a correct answer but lacks confidence in its correctness, subsequently producing a wrong answer after self-correction.
\item \textbf{Critical} (\XSolidBrush $\rightarrow$ \Checkmark): The model initially generates a wrong answer but arrives at a correct answer through effective reflection.
\item \textbf{Stubborn} (\XSolidBrush $\rightarrow$ \XSolidBrush): The model initially generates a wrong answer and stubbornly insists on this incorrect answer.
\end{enumerate}

Essentially, model confidence in correct answers (case 1) and lack of confidence (case 2) are inversely related; likewise, the reflection capacity (case 3) and obstinacy in incorrect answers (case 4) are also inversely equivalent. Thus, the four self-correction cases can be distilled into two key capacities: \textbf{Confidence Capability} (confidence in correct answers) and \textbf{Critique Capability} (the ability to correct wrong answers). 


\section{Evaluation Metrics}

\label{sec:methodology}

To further investigate the two decomposed capabilities in \S \ref{sec:decomposition}, we first formalize the problem and introduce relevant mathematical notations (\S \ref{subsec:notations}). Then we propose two metrics from a probabilistic perspective to measure these two capabilities, and demonstrate that model performance after self-correction (i.e., accuracy) is essentially a weighted sum of these two metrics (\S \ref{subsec:confidence}). 
Also, a unified metric to measure overall self-correction capability is proposed in \S \ref{subsec:RSS}.
Since the computation of our metrics depends on the probability of events, we then provide probability estimation methods in Appendix \ref{app:probability_estimation} and analyze metric convergence in Appendix \ref{app:convergence}. 

\subsection{Problem Formulation and Notations}
\label{subsec:notations}



Initially, we have a set comprising of $n$ questions, denoted as $A = \{ q_1, q_2, ..., q_n \}$. For a given question $q_i$, the probability of the model generates a correct answer through a single temperature-based sampling before and after self-correction are denoted as $P(a_i)$ and $P(b_i)$, respectively. 
We define a stochastic process: 

\textbullet~ Randomly sampling a question $q$ from $A$ with equal probability.

In the above random process, the probability of the model generating a correct answer for the question $q$ before and after self-correction is denoted as $P(a)$ and $P(b)$, respectively. We define their expectations as $Accuracy_1$ and $Accuracy_2$ ($Acc_1$ and $Acc_2$ for short), then we have:

\begin{equation}
Acc_1 = E[P(a)] = \frac{\sum_{i=1,...,n}P(a_i)}{n} 
\end{equation}

\begin{equation}
Acc_2 = E[P(b)] = \frac{\sum_{i=1,...,n}P(b_i)}{n} 
\end{equation}
For convenience, all of the notations mentioned and their meanings are shown in Appendix \ref{app:notations}.

\subsection{Confidence Level and Critique Score}
\label{subsec:confidence}

\begin{figure}[!tb]
    \centering
    \includegraphics[width=0.48\textwidth]{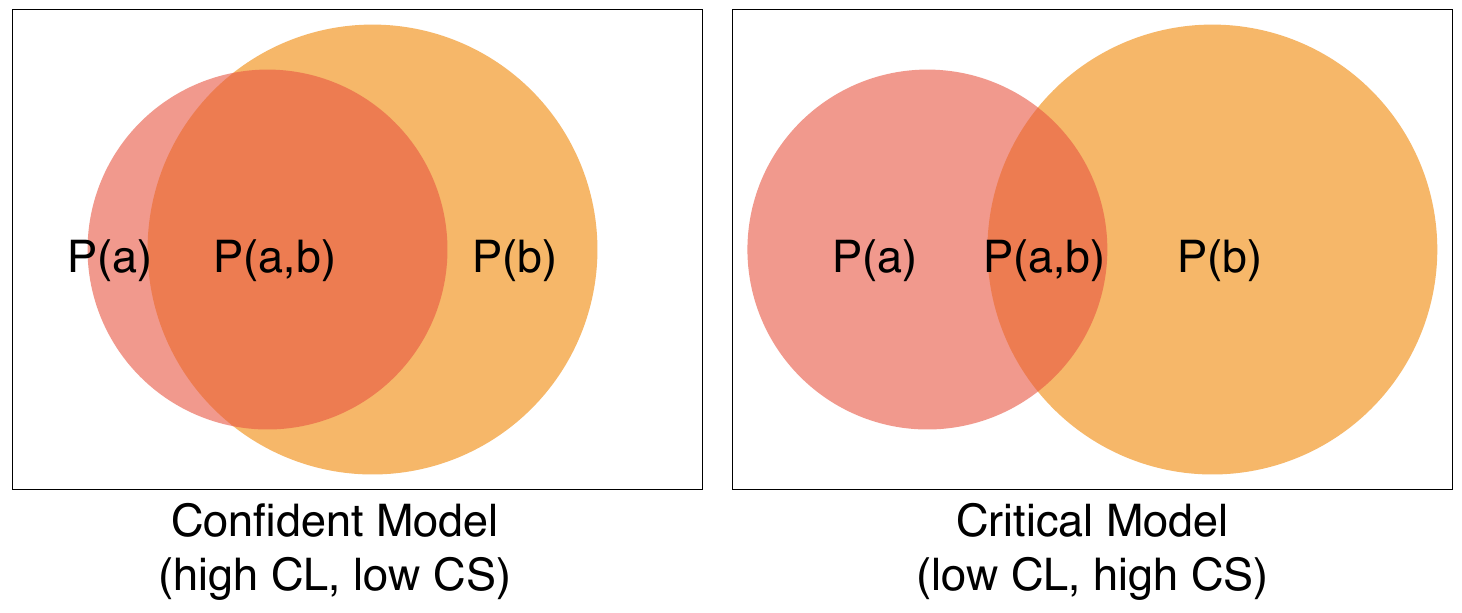}
    \caption{Venn diagram for confident/critique models in complete probability space. The \colorbox{venn_red}{red}, \colorbox{venn_orange}{orange} circles and their overlap area denote the probability of a model correctly answering questions before self-correction, after self-correction, and both respectively. the overlap area of confident models is much larger than that of critical models.}
    \label{fig:Venn}
\end{figure}

How confident are LLMs in their correct answers? To answer this question from a probabilistic perspective, we introduce a metric named \textbf{C}onfidence \textbf{L}evel (\textbf{CL}). Similarly, to measure the capability to critique and turn wrong answers to correct, we introduce another metric termed \textbf{C}ritique \textbf{S}core (\textbf{CS}). CL/CS is defined as the conditional probability of a model generating a correct answer after self-correction given it has generated a correct/wrong one initially, then we have:

\begin{equation}
\label{equ:CL}
CL = E[P(b|a)] 
= \frac{\sum_{i=1}^{n}P(a_i)P(b_i|a_i)} {\sum_{i=1}^{n}P(a_i)},
\end{equation}
\begin{equation}
CS = E[P(b|\neg a)] 
= \frac{\sum_{i=1}^{n}[1-P(a_i)]P(b_i|\neg a_i)}{\sum_{i=1}^{n}[1-P(a_i)]},
\end{equation}

where $P(b_i|a_i)/P(b_i|\neg a_i)$ is the conditional probability of a model correctly answering $q_i$ after self-correction given that it has answered it correctly/wrong initially, and the derivation details are shown in Appendix \ref{app:metric_derivation}.
To intuitively illustrate CL/CS, we present a Venn diagram in Figure \ref{fig:Venn} to compare two types of models.

Intuitively, a model with a strong self-correction ability tends to show a higher \( Acc_2 \), which is caused by its high CL and CS. We also find the accuracy after self-correction ($Acc_2$) satisfies the following relationship (with derivation shown in Appendix \ref{subapp:weighted_sum_proof}):
\begin{equation}
\label{equ:weighted_sum}
Acc_2 =Acc_1*CL+(1-Acc_1)*CS
\end{equation}
Essentially, $Acc_2$ is the weighted sum of $CL$ and $CS$ with weights $Acc_1$ and $1 - Acc_1$ respectively, and improving CL/CS will increase $Acc_2$. Besides, this equation also further validates the rationality of our decomposition in \S \ref{sec:decomposition}.

\subsection{Relative Self-Correction Score}
\label{subsec:RSS}

Measuring self-correction capability with a single unified metric.
The above two metrics respectively reflect different aspects of self-correction capability, which is beneficial for a detailed analysis. 
However, it is hard to compare the overall self-correction ability of two models with these two metrics, as one model may process a higher CL while the other exhibits a higher CS. Another potential metric that can reflect self-correction capability is $Acc_2$, but it can be significantly influenced by the initial ability (i.e. $Acc_1$). 
For instance, in \S \ref{sec:experiments} Llama3-8B-Instruct shows an $Acc_1$ of 71.0\% and an $Acc_2$ of 78.1\% on the GSM8k, indicating a substantial improvement in accuracy after self-correction. Conversely, GPT-4 Turbo has an $Acc_1$ of 93.6\% and an $Acc_2$ of 92.1\%, showing a slight decrease in accuracy. 
Intuitively, Llama3-8B-Instruct seems to possess better self-correction ability, yet GPT-4 Turbo has a higher $Acc_2$.

\begin{figure}[!tb]
    \centering
    \includegraphics[width=0.48\textwidth]{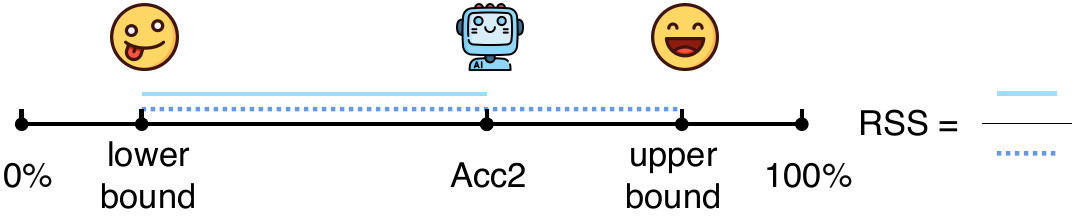}
    \caption{Visualized expression of Relative Self-correction Score.}
    \label{fig:RSS} 
\end{figure}

To fairly compare the overall self-correction capabilities of different models and eliminate the influence of $Acc_1$, we propose the \textbf{R}elative \textbf{S}elf-Correction \textbf{S}core (\textbf{RSS}), which is essentially a normalized form of $Acc_2$. Similar to \citet{yang-etal-2024-large-language-models-always}, we derive the upper and lower bounds of $Acc_2$ and define RSS as the position of the actual $Acc_2$ within this range (also shown in Figure \ref{fig:RSS}):
\begin{equation}
RSS = \frac{Acc_2 - Acc_2^{low}}{Acc_2^{upp} - Acc_2^{low}} = \frac{Acc_2 - Acc_1^2}{2Acc_1-2Acc_1^2},
\end{equation}

where $Acc_2^{low} = Acc_1^2, Acc_2^{upp} =2Acc_1 - Acc_1^2$ denotes lower and upper bound of $Acc_2$ respectively, with derivation details shown in Appendix \ref{app:RSS_derivation}.
Empirically we have $RSS \in (0,1)$, and higher $RSS$ indicates better self-correction capability.
Specifically, when there is no change in accuracy after self-correction (i.e. $Acc_1$=$Acc_2$), we have $RSS = 0.5$. 
$RSS > 0.5$ signifies an increase in accuracy after self-correction, whereas $RSS < 0.5$ indicates a decrease.

\section{Experiments}
\label{sec:experiments}
\begin{table*}[!tb]
  \centering
  \footnotesize
    \begin{tabular}{l|cccc|cccc|cccc}
    \toprule
          \multicolumn{1}{c|}{\multirow{2}[2]{*}{Models}} & \multicolumn{4}{c|}{GSM8k}    & \multicolumn{4}{c|}{MMLU}     & \multicolumn{4}{c}{BoolQ} \\
    & $Acc_1$  & $Acc_2$  & $CL$    & $CS$    & $Acc_1$  & $Acc_2$  & $CL$    & $CS$    & $Acc_1$  & $Acc_2$  & $CL$   & $CS$  \\
    \midrule
    Llama3-8B-Instruct & 71.0  & 78.1  & 91.7  & \textbf{44.9} & 62.2  & 64.0  & 94.9  & 13.1  & 62.3  & 64.8  & 86.0  & 29.8    \\
    Deepseek-7B-Chat & 61.2  & 60.9  & 95.9  & 5.6   & 47.8  & 47.9  & \textbf{98.7} & 1.3   & 57.8  & 57.6  & \textbf{98.8} & 1.2   \\
    Mistral-7B-Instruct & 50.1  & 51.1  & 90.9  & 11.0  & 59.2  & 59.2  & 98.4  & 2.3   & 61.4  & 62.5  & 98.5  & 5.4  \\
    Qwen2.5-7B-Chat & 91.9  & 92.4  & \textbf{99.4} & 14.5  & 71.0  & 71.5  & 93.3  & 18.0  & 58.8  & 60.9  & \textbf{93.9}  & 13.8  \\
    GLM4-9B-Chat & 64.9  & 63.7  & 87.9  & 19.0  & 63.5  & 64.6  & 83.3  & \textbf{32.1} & 61.1  & 64.8  & 77.1  & \textbf{45.5}  \\
    \midrule
    Llama3-70B-Instruct & 90.7  & 92.7  & 97.3  & \textbf{48.1} & 78.2  & 79.5  & 97.2  & \textbf{16.2} & 76.3  & 76.4  & 84.7  & 49.3  \\
    Deepseek-67B-Chat & 82.4  & 82.3  & 99.1  & 3.7   & 65.3  & 66.3  & 94.8  & 12.9  & 69.8  & 69.8  & 89.9  & 23.4  \\
    Qwen2.5-72B-Chat & 95.7  & 95.9  & \textbf{99.9} & 7.5   & 82.6  & 83.4  & \textbf{98.2} & 13.5  & 65.5  & 75.9  & 93.9  & 41.5 \\
    \midrule
    Qwen-Max & 96.1  & 96.4  & \textbf{99.9} & 11.5  & 83.8  & 85.0  & \textbf{99.2} & 11.6  & 71.3  & 73.6  & 98.2  & 12.5  \\
    GPT-3.5 Turbo  & 81.3  & 84.0  & 95.6  & \textbf{33.8} & 65.3  & 65.6  & 89.6  & 20.5  & 68.5  & 70.3  & 75.7  & \textbf{58.8}  \\
    GPT-4 Turbo  & 93.6  & 92.1  & 96.8  & 23.9  & 84.3  & 82.3  & 88.4  & \textbf{49.6} & 80.5  & 78.6  & 87.8  & 40.6  \\
    \bottomrule
    \end{tabular}%
\caption{Experiment results on GSM8k, MMLU and BoolQ. We report accuracy(\%) before and after self-correction (denoted as $Acc_1$ and $Acc_2$). Confidence Level ($CL$) and Critique Score ($CS$) are also shown for fine-grained analysis of self-correction behavior.}
\label{tab:main_results}%
\end{table*}%
\subsection{Experimental Setup}

\paragraph{Models}
Experiments are conducted on both open-source and closed-source models. For the closed-source models, we assess Qwen-Max \citep{qwen}, GPT-3.5 Turbo, and GPT-4 Turbo \citep{gpt4} by API calls. For the open-source models, we evaluate Llama3-(8B,70B) \citep{llama3}, Qwen2.5-(7B,72B) \citep{yang2024qwen2}, DeepSeek-LLM-7B \citep{deepseek-llm}, Mistral-7B-v3 \citep{mistral}, and GLM4-9B \citep{glm2024chatglm}, and parameters of these models are publicly available on HuggingFace \footnote{\url{https://huggingface.co/}}.

\paragraph{Dataset}
We evaluate self-correction capability on both classification and generation tasks, including domains in mathematics, coding, instruction following, common-sense reasoning, and knowledge. To be specific, the dataset we utilized include GSM8k \citep{cobbe2021gsm8k}, Humaneval \citep{chen2021codex}, IFEval \citep{zhou2023instruction}, MMLU \citep{hendrycks2021measuring}, BoolQ \citep{clark-etal-2019-boolq}, and CommonsenseQA \citep{talmor-etal-2019-commonsenseqa}.

More implementation details are shown in Appendix \ref{subapp:setup1}.

\subsection{Experimental Results}

\begin{figure}[!tb]
    \centering
    \includegraphics[width=0.49\textwidth]{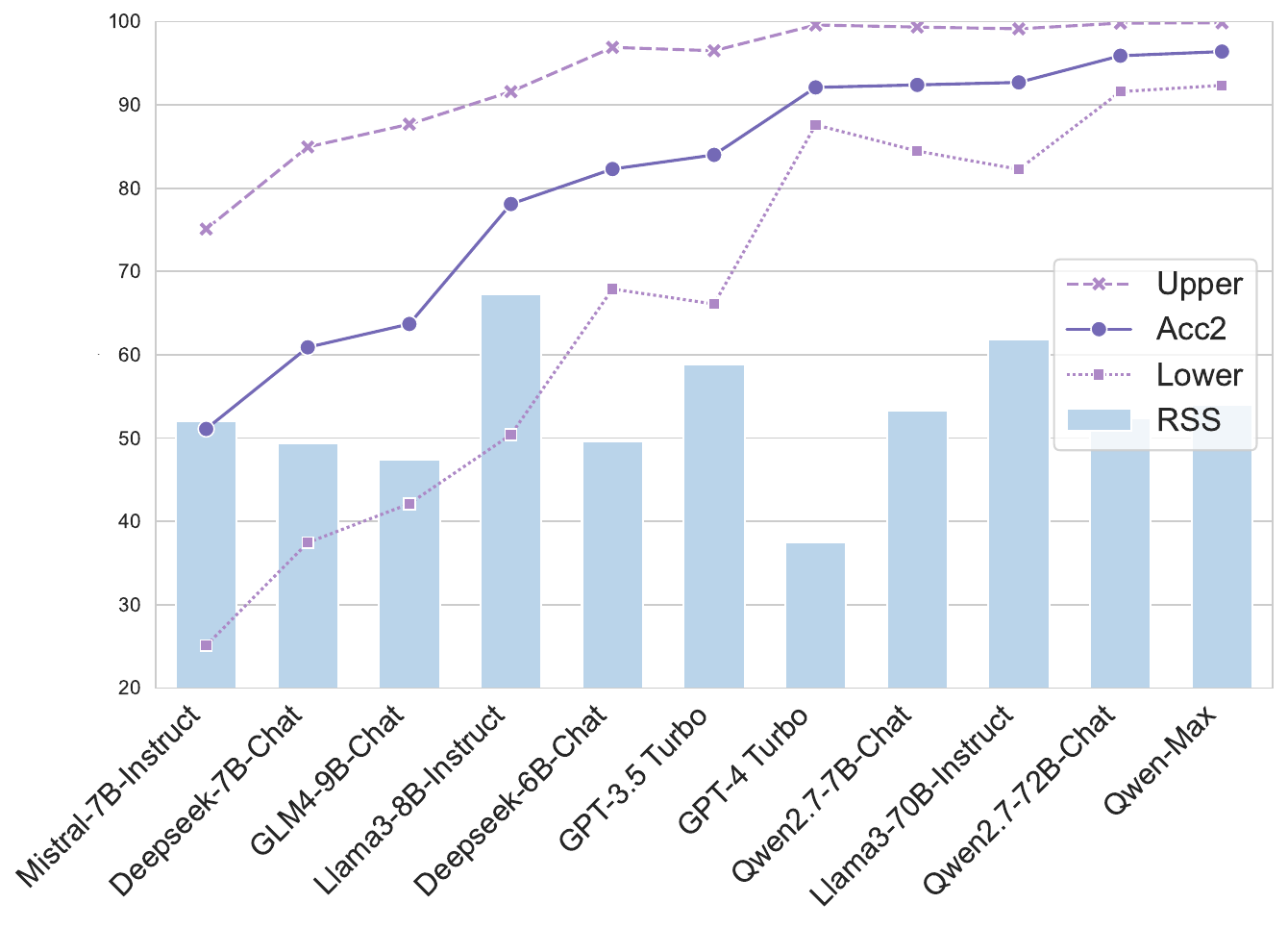}
    \caption{Relative Self-correction Score (RSS) results on GSM8k (shown in ascending order of $Acc_2$). Except for showing RSS for each evaluated model in a bar, we also show $Acc_2$, upper and lower bounds of $Acc_2$ in lines of different colors for comparison. }
    \label{fig:RSS_results}
\end{figure}
Self-correction capability evaluation experiments are conducted on various models and Accuracy (\%) before and after self-correction is reported. We also report Confidence Level and Critique Score during the self-correction process for fine-grained analysis, as the results shown in Table \ref{tab:main_results} and \ref{tab:main_results_more}.
To measure overall self-correction capability and remove the effect of initial Accuracy, we show Relative Self-correction Score results on GSM8k in Figure \ref{fig:RSS_results}, and more results are illustrated in Table \ref{tab:RSS_results}.
Our findings include:

1. \textit{Self-correction does not necessarily lead to an increase in Accuracy.} For example, on the GSM8k dataset, accuracy of GPT-3.5 Turbo is improved by 2.7\% after self-correction, whereas accuracy of GPT-4 Turbo is decreased by 1.5\%. 
As a result, RSS of GPT-3.5 Turbo is much higher than that of GPT-4 Turbo.

2. \textit{In general, the CL values are relatively high, while the CS values are relatively low.} This indicates that models tend to have high confidence but still have considerable room for improvement in their critique capabilities. Furthermore, models with higher CS values (e.g., Llama3-8-Instruct) tend to process lower CL values, suggesting that it may be hard for models to achieve both high confidence and critique capabilities simultaneously.

3. \textit{Different models exhibit distinct behaviors.} For instance, Deepseek-7B-Chat and Mistral-7B-Instruct are generally more "conservative", tending not to alter their answers after self-correction, resulting in high CL and low CS. On the other hand, Llama3-8B-Instruct and GLM4-9B-Chat are more "liberal", often overturning their initial answers and providing new ones after self-correction, which leads to low CL and high CS.

4. \textit{Models from the same series tend to show similar behaviors.} For example, both Llama3-8B-Instruct and Llama3-70B-Instruct exhibit low CL and high CS, whereas Qwen2.5-7B-Chat and Qwen2.5-72B-Chat tend to show high CL and low CS, and this phenomenon indicates confidence and critique capabilities are likely influenced by the pre-training data.

\section{Behavior Manipulation}

\begin{table*}[htbp]
  \centering
  \footnotesize
    \begin{tabular}{c|ll|ll|ll|ll}
    \toprule
     \multirow{2}[1]{*}{Prompt} & \multicolumn{2}{c|}{GSM8k} & \multicolumn{2}{c|}{MMLU} & \multicolumn{2}{c|}{BoolQ} & \multicolumn{1}{c}{Avg}   & \multicolumn{1}{c}{Avg} \\
          & \multicolumn{1}{c}{$CL$}    & \multicolumn{1}{c|}{$CS$}    & \multicolumn{1}{c}{$CL$}    & \multicolumn{1}{c|}{$CS$}    & \multicolumn{1}{c}{$CL$}    & \multicolumn{1}{c|}{$CS$}   & \multicolumn{1}{c}{$CL$}    & \multicolumn{1}{c}{$CS$} \\
    \midrule
      Reask & $91.7_{0.0}$  & $44.9_{0.0}$  & $94.9_{0.0}$  & $13.1_{0.0}$  & $86.0_{0.0}$  & $29.8_{0.0}$  & $90.9_{0.0}$  & $29.3_{0.0}$  \\
          Confidence & $93.5_{\textcolor{red}{+1.8}}$  & $32.9_{\textcolor{green}{-12.0}}$  & $99.0_{\textcolor{red}{+4.1}}$  & $2.0_{\textcolor{green}{-11.1}}$  & $96.1_{\textcolor{red}{+10.1}}$  & $8.9_{\textcolor{green}{-20.9}}$  & $96.2_{\textcolor{red}{+5.3}}$  & $14.6_{\textcolor{green}{-14.7}}$  \\
          Critique & $77.7_{\textcolor{green}{-14.0}}$  & $47.9_{\textcolor{red}{+3.0}}$  & $71.1_{\textcolor{green}{-23.8}}$  & $26.0_{\textcolor{red}{+22.9}}$  & $54.6_{\textcolor{green}{-31.4}}$  & $62.3_{\textcolor{red}{+32.5}}$  & $67.8_{\textcolor{green}{-23.1}}$  & $48.7_{\textcolor{red}{+19.4}}$  \\
    \bottomrule
    \end{tabular}%
    \caption{Self-correction behavior under different kinds of prompts.
    \textcolor{green}{Green} and \textcolor{red}{red} text denotes the change in accuracy of "Confidence"/"Critique" prompt relative to "Reask" prompt baseline.}
  \label{tab:prompt}%
\end{table*}%



\begin{figure}[!tb]
    \centering
    \includegraphics[width=0.48\textwidth]{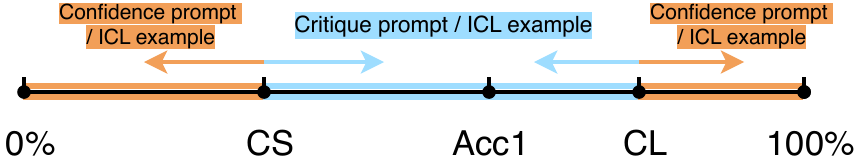}
    \caption{A trade-off between CL and CS. Confidence prompt/ICL example can lead higer CL and lower CS; critique prompt/ICL example can cause lower CL and higher CS.}
    \label{fig:trade-off}
\end{figure}

\label{sec:Analysis}
In this section, we explore manipulating self-correction behavior of LLMs without fine-tuning. 
We try to utilize different prompts (\S \ref{subsec:manipulation_by_prompt}), provide different in-context learning (ICL) examples (\S \ref{subsec:manipulation_by_ICL}), and observe the change in self-correction behavior.
Experimental results indicate it is hard to consistently enhance both confidence and critique capabilities simultaneously through prompt or ICL, and we also illustrate the trade-off between CL and CS in Figure \ref{fig:trade-off}.
Improving one aspect often leads to a decline in the other, 
so there is no guarantee of improving overall self-correction capability simply by different prompts or ICL examples.


\subsection{Manipulation by Prompt}
\label{subsec:manipulation_by_prompt}
In \S \ref{sec:experiments}, our prompt to encourage LLMs to self-correct is simply to ask LLMs the question again. By taking this as a baseline, we try two other prompt strategies and make a comparison.
\citet{huanglarge} utilizes a critique prompt to encourage LLMs to find errors in answers, while \citet{li2024confidence} emphasizes the importance of confidence in correct answers. Inspired by previous research, we attempt confidence prompt and critique prompt to manipulate the self-correction behavior of Llama3-8B-Instruct (see Appendix \ref{app:prompt} for prompt details), with experimental results presented in Table \ref{tab:prompt}. We observe that confidence prompt enhances CL across all tasks but diminishes CS. Conversely, critique prompt improves CS but the price is a reduction in CL. 
To improve self-correction capability of LLM, we should improve both confidence and critique simultaneously, which can be hardly achieved by simply changing a different prompt.
Besides, the debate (\S \ref{sec:intrudcution}) on whether self-correction can improve performance can also caused by the difference in prompts.

\begin{figure}[!t]
    \centering
    \includegraphics[width=0.45\textwidth]{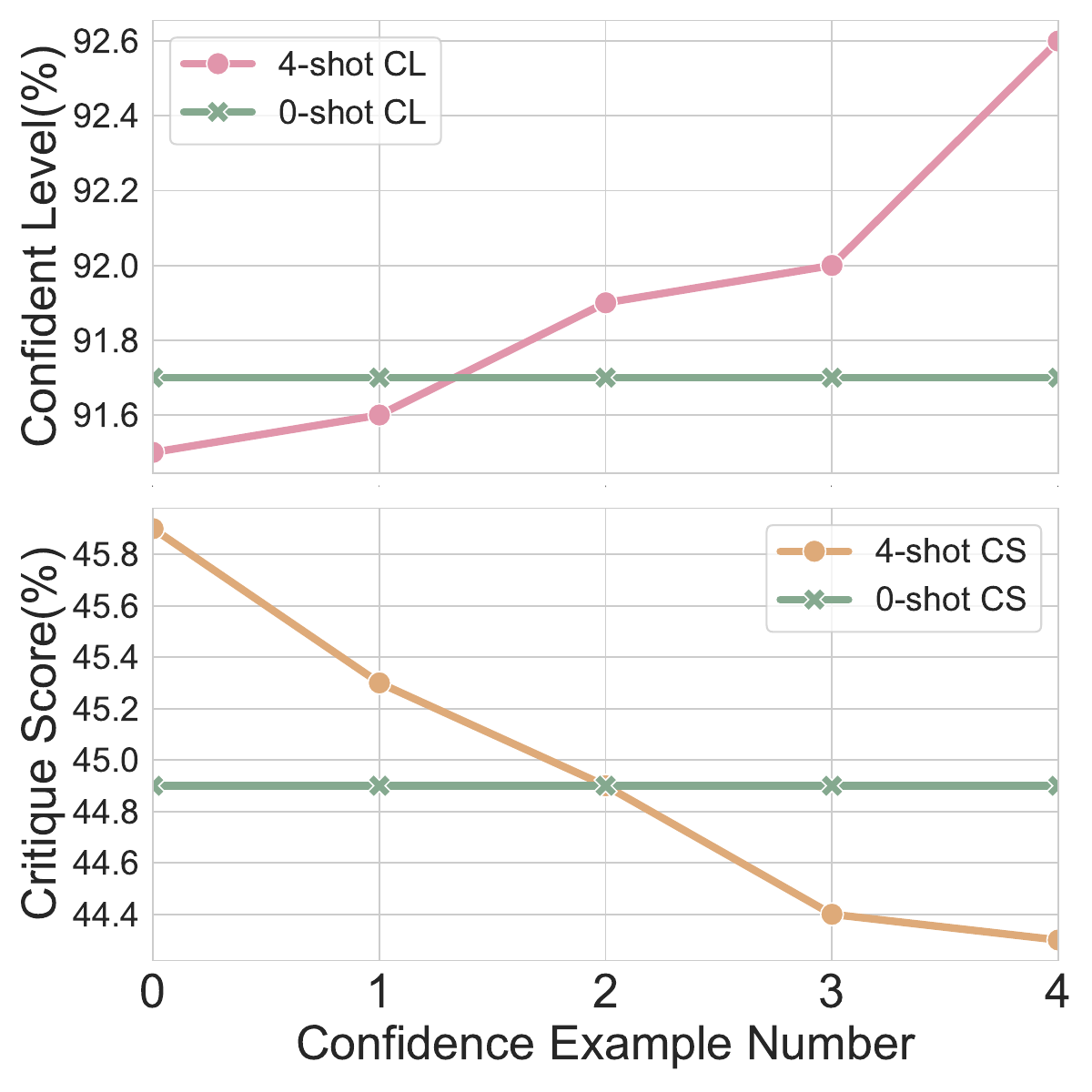}
    \caption{Self-correction behavior of 4-shot ICL with different confidence example numbers. With the increase of confident example number, $CL$ increases, and $CS$ decreases.}
    \label{fig:icl_gsm8k}
\end{figure}
\subsection{Manipulation by ICL}
\label{subsec:manipulation_by_ICL}
Prior work \citep{dong-etal-2024-survey,yang-etal-2023-demonstration} has demonstrated that LLMs can do in-context learning by providing only a few examples, and we explore manipulating self-correction by ICL examples in the form of case 1 (confidence example) and case 3 (critique example) in \S \ref{sec:decomposition}.
In confidence example, model generates a correct answer and maintains it after self-correction; while in critique example, model gives a wrong answer but successfully corrects it after self-correction. 
We evaluate the Llama3-8B-Instruct model under a 4-shot setting and utilize the 0-shot setting as a baseline for comparison, varying the number of confidence and critique examples among the four examples used.
As the experimental results shown in Figure \ref{fig:icl_gsm8k}, we find that a higher number of confidence examples increases confidence but diminishes critique capability, whereas more critique examples enhance CS but reduce CL. When the number of these two examples is the same (2:2), model behavior is similar to that of 0-shot setting.

\section{Improvement Tuning}



\label{sec:tuning}
We have decomposed self-correction capability into confidence capability and critique capability (\S \ref{sec:decomposition}) and find a trade-off between them without fine-tuning (\S \ref{sec:Analysis}). In this section, we further explore training models to acquire better self-correction performance by improving both the above two capabilities simultaneously, and propose a fine-tuning method named Confidence-and-Critique Improvement Tuning (CCT),
which can be divided into \textbf{C}onfidence \textbf{L}evel Improvement \textbf{T}uning (\textbf{CLT}) and \textbf{C}ritique \textbf{S}core Improvement \textbf{T}uning (\textbf{CST}). CLT is designed to increase confidence capability, while CST aims to enhance the critique capacity. 

\paragraph{A theoretical comparison of different training methods.} Vanilla Supervised Fine-Tuning (SFT) teaches the model how to complete a task (i.e. how to generate the correct answer for a given question), but this paradigm can hardly teach a model how to reflect and self-correct.
In contrast, CLT provides a user question and a correct answer as the context, training the model to be confident in this correct answer. Similarly, CST gives a user question accompanied by a wrong answer as the context and teaches model critique capability by taking a correct answer as supervision.
CLT and CST training data can be acquired by automatic transformation of SFT training set, and an example of these training data is shown in Appendix \ref{app:example_training-data}.
CCT training data is essentially a mixture of CLT and CST, improving self-correction by combining the advantages of them. 
There are also other self-correction improvement training methods \citep{yan2024s,han2024small,welleck2023generating} 
with strong verifiers \citep{zhang-etal-2024-small,chen2024teaching} or reinforcement learning \citep{kumar2024training}, but CCT is much simpler and can be achieved by automatic transformation of SFT data, so we do not compare CCT to these methods and only investigate the improvement to SFT.

\begin{table}[!t]
  \footnotesize
  \centering
    \begin{tabular}{cc|cccc}
    \toprule
    Task  & Method & $Acc_1$  & $Acc_2$  & $CL$    & $CS$ \\
    \midrule
    \multirow{4}[2]{*}{GSM8k} & SFT   & \textbf{39.3 } & 40.3  & 75.2  & 17.7  \\
          & CLT & 30.3  & 34.2  & 94.6  & 8.0  \\
          & CST & 33.1  & 42.2  & 80.5  & 23.2  \\
          & CCT & 36.0  & \textbf{44.2 } & 89.9  & 18.4  \\
    \midrule
    \multirow{4}[2]{*}{MMLU} & SFT   & 48.6  & 48.9  & 70.3  & 28.6  \\
          & CLT & 26.4  & 26.4  & 99.9  & 0.1  \\
          & CST & 47.6  & 27.4  & 5.1   & 47.6  \\
          & CCT & \textbf{51.2}  & \textbf{55.5 } & 85.5  & 24.0  \\
    \midrule
    \multirow{4}[2]{*}{BoolQ} & SFT   & \textbf{63.6 } & 63.8  & 75.8  & 42.8  \\
          & CLT & 53.8  & 53.8  & 99.1  & 1.0  \\
          & CST & 58.8  & 41.5  & 1.3   & 98.9  \\
          & CCT & 62.4  & \textbf{74.0 } & 83.7  & 57.8  \\
    \bottomrule
    \end{tabular}%
    \caption{Experiment results of different training methods on GSM8k, MMLU and BoolQ. CCT outperforms SFT in $Acc_2, CL, CS$, showing better self-correction capability, and we also show results for CLT and CST for comparison.}
  \label{tab:CCT}%
\end{table}%

\begin{figure}[!tb]
    \centering
    \includegraphics[width=0.49\textwidth]{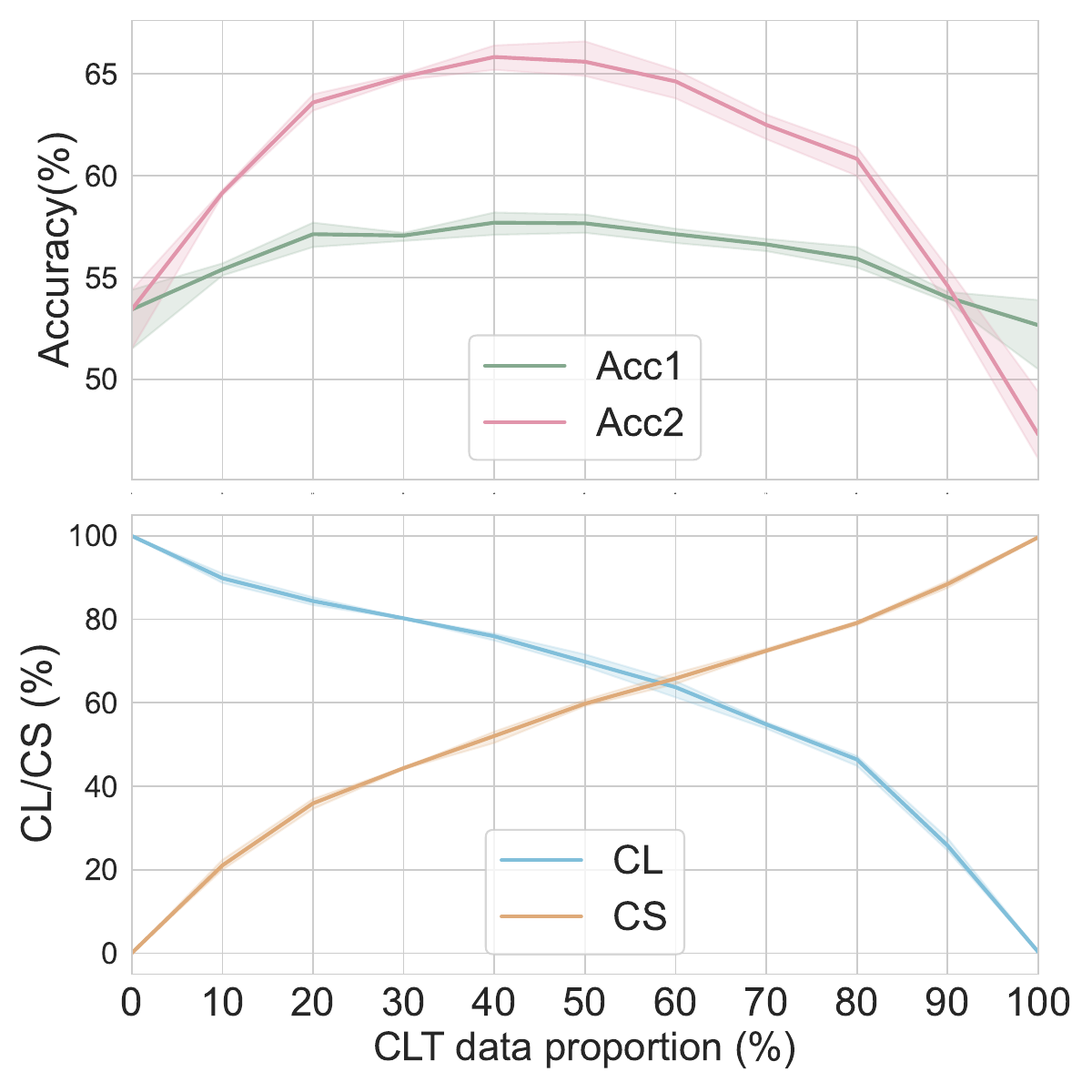}
    \caption{Self-correction behaviors under different proportions of CLT and CST training data on BoolQ.}
    \label{fig:CCT_prop}
\end{figure}

\paragraph{An empirical comparison of these training methods.} 
We fine-tune Llama2-7B-Base on three tasks by the above training approaches with Lora \citep{hu2021lora}, and more implementation details are shown in Appendix \ref{subapp:setup2}. 
As the experimental results displayed in Table \ref{tab:CCT}, we report Accuracy (\%) before and after self-correction (denoted as $Acc_1, Acc_2$) of fine-tuned models under different training strategies, along with $CL$ and $CS$ for fine-grained analysis. 
Our findings indicate that while SFT achieves the best initial performance ($Acc_1$), it exhibits relatively weak self-correction capability and achieves minimal performance improvement after self-correction. 
On the other hand, CLT and CST significantly enhance confidence and correction abilities, respectively, yielding the highest CL or CS. 
However, these single-focus tuning strategies often substantially compromise model capability in the other aspect, even leading to negative performance gains after self-correction. 
In contrast, CCT can enhance both confidence and critique capabilities simultaneously, and the corresponding CL and CS generally surpass those of SFT. 
Notably, CCT can lead to considerable accuracy improvements after self-correction and achieve the highest $Acc_2$ across all three tasks, significantly outperforming other methods, which suggests that CCT can effectively enhance the self-correction capabilities of LLMs.


\paragraph{Exploring the proportions of CLT and CST.}
Empirical results have shown a single CLT or CST can not improve self-correction capability, but a mixture of them (CCT) can be effective. 
We further investigate performance of fine-tuned models under different mixing ratios by keeping the total size of the training set constant while adjusting the proportions of the two types of data.
We test each data mixture three times with different random seeds and report the average result, as the experimental results on BoolQ shown in Figure \ref{fig:CCT_prop}. 
We find that as the proportion of CLT data increases, CL consistently rises, while the CS value monotonically decreases. $Acc_1$ and $Acc_2$ exhibit an inverted U-shaped curve (initially increasing and then decreasing), and the model achieves its highest self-correction performance when the proportion of CLT data is approximately 40\%.

\begin{figure}[!tb]
    \centering
    \includegraphics[width=0.49\textwidth]{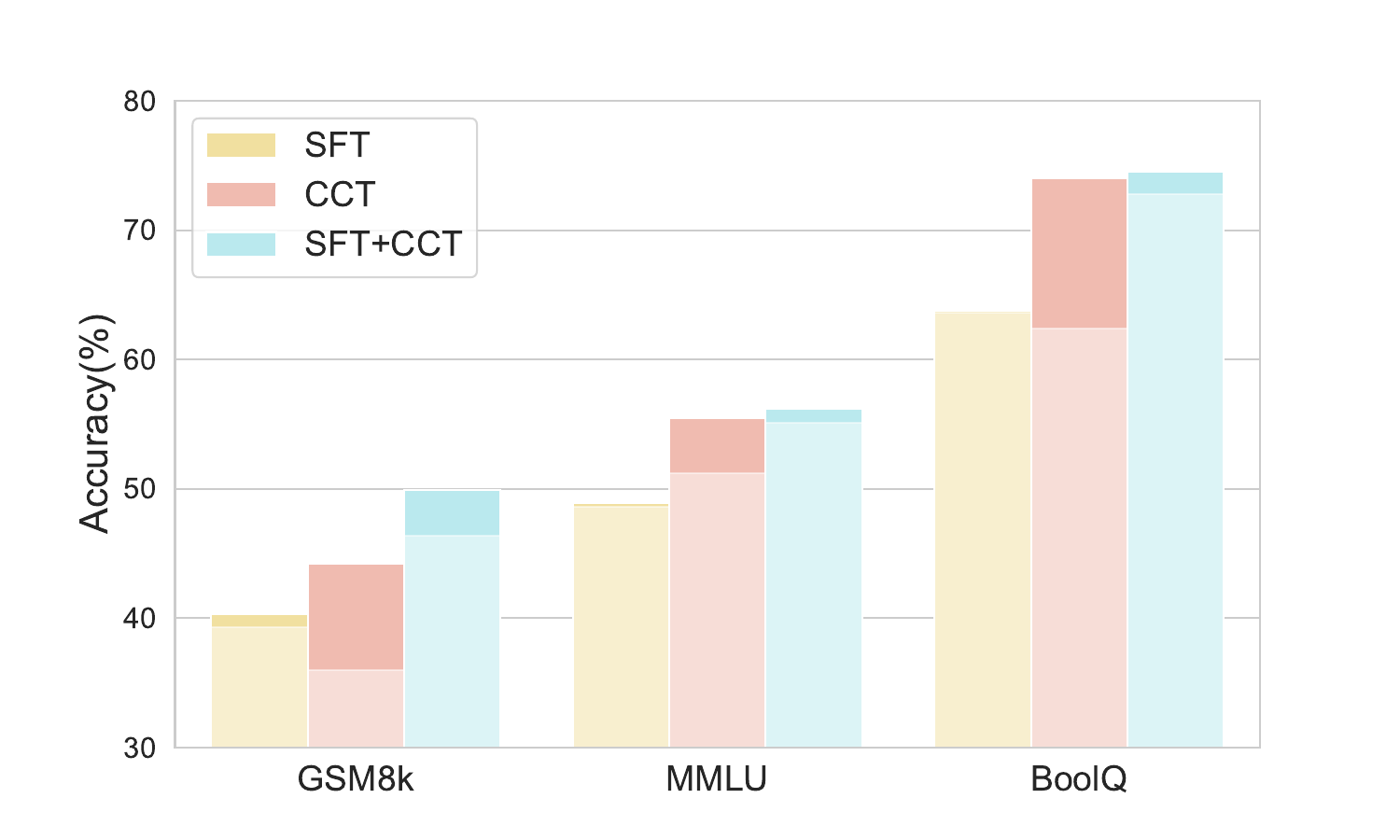}
    \caption{A comparison of SFT, CCT, and SFT+CCT. $Acc_2$ is presented in colorful bars and the whited parts denote $Acc_1$. SFT+CCT can achieve both high $Acc_1$ and $Acc_2$.}
    \label{fig:bar}
\end{figure}


\paragraph{Can we combine CCT with SFT?} Since SFT can make model achieve high $Acc_1$ and CCT achieves high $Acc_2$, we then explore combining them for both high $Acc_1$ and $Acc_2$. As the results shown in Figure \ref{fig:bar}, SFT achieves high $Acc_1$ but low $Acc_2$; CCT achieves high $Acc_2$ but $Acc_1$ is relatively low; and SFT+CCT can achieve both high $Acc_1$ and $Acc_2$. This phenomenon indicates that we can improve self-correction capability in SFT stage by adding some CCT data. Since CCT data can be acquired from SFT data, we can also treat CCT as an effective data augmentation strategy.

\section{Related Work}
\paragraph{Self-Correction}

LLMs can correct responses by themselves \citep{liu2024large} or with external feedback \citep{jiang2023selfevolve}, and this self-correction capability can be improved by prompting \citep{li2024confidence,wu-etal-2024-large} or fine-tuning \citep{welleck2023generating,kumar2024training}. 
Unlike previous work, we provide a new perspective to decompose, evaluate, analyze, and improve self-correction. 

\paragraph{Evaluation and Metrics} 
The evaluation of LLMs \citep{chang2023surveyevaluationlargelanguage} mainly focuses on specific capabilities (e.g. mathematics \citep{gao2024omni}, instruction-follow \citep{zhou2023instruction}) or properties (e.g. MBTI \citep{pan2023llmspossesspersonalitymaking}, consistency \citep{yang-etal-2024-large-language-models-always}). 
We evaluate self-correction capability with metrics derived from a probabilistic perspective.

\paragraph{Post-Training}
LLMs usually require further post-training to enhance specific capabilities after pre-training. SFT \citep{zhang2023instruction,weifinetuned} can improve general ability on multiple tasks; 
RLHF \citep{ouyang2022traininglanguagemodelsfollow} and DPO \citep{rafailov2024directpreferenceoptimizationlanguage,gao2024unifiedviewpreferencelearning} can align LLMs with human preference. 
Our CCT improves self-correction capability by transforming the format of SFT data and be combined with SFT.


\section{Conclusion}

We propose a methodology to decompose, evaluate, and analyze the self-correction capabilities of LLMs.
By enumerating four cases, we decompose self-correction capability into confidence capability and critique capability, and propose two metrics from a probabilistic perspective to measure these two capabilities, along with another metric to measure the overall self-correction capability.
Based on our metrics and probability estimation methods, we conduct extensive experiments and draw some empirical conclusions. A trade-off between these two capabilities is also observed when manipulating behaviors by prompt or ICL, 
and further we propose a simple yet efficient training strategy for self-correction improvement by transforming data format in SFT stage.
To summarize, our decomposition and evaluation methodology can be helpful to self-correction behavior analysis and our training strategy can improve self-correction capability, thus paving the way for further exploration in LLM self-correction.

\clearpage

\section*{Limitations}

The calculation of our proposed metrics relies on probability estimation, which necessitates repeated sampling for the same question, being more computationally expensive than traditional non-probability evaluation.

Our decomposition and analysis are simplified and real self-correction can be more complex.
For instance, generating wrong answers before and after self-correction might be due to 1. the model stubbornly adhering to an incorrect answer or 2. the question being too hard and beyond current capability of the model.
Our analytical approach can not distinguish between these two scenarios and treats them the same.
Besides, our evaluation methodology can only reflect the self-correction capability on a whole dataset, but can not indicate which type of questions is more likely to cause the model to exhibit confidence or critique behaviors, and identifying these questions for a given model still requires human efforts in case studies.
Thus, we leave a more detailed and fine-grained analysis of self-correction to future work.

Although we have observed that models from the same series exhibit similar self-correction behaviors and hypothesize that these behaviors are influenced by the pre-training data, the underlying reasons for how these behaviors come into being remain unknown, and we leave further explorations on deeper reasons to further work.

\section*{Ethical Considerations}

The data we utilized are open for research, and evaluated LLMs are all publicly available by either parameters or API calls. 
Therefore, we do not anticipate any ethical concerns in our research.


\bibliography{anthology,custom}

\begin{thebibliography}{50}
\providecommand{\natexlab}[1]{#1}

\bibitem[{Achiam et~al.(2023)Achiam, Adler, Agarwal, Ahmad, Akkaya, Aleman, Almeida, Altenschmidt, Altman, Anadkat et~al.}]{gpt4}
Josh Achiam, Steven Adler, Sandhini Agarwal, Lama Ahmad, Ilge Akkaya, Florencia~Leoni Aleman, Diogo Almeida, Janko Altenschmidt, Sam Altman, Shyamal Anadkat, et~al. 2023.
\newblock Gpt-4 technical report.
\newblock \emph{arXiv preprint arXiv:2303.08774}.

\bibitem[{AI@Meta(2024)}]{llama3}
AI@Meta. 2024.
\newblock \href {https://github.com/meta-llama/llama3/blob/main/MODEL_CARD.md} {Llama 3 model card}.

\bibitem[{Bai et~al.(2023)Bai, Bai, Chu, Cui, Dang, Deng, Fan, Ge, Han, Huang et~al.}]{qwen}
Jinze Bai, Shuai Bai, Yunfei Chu, Zeyu Cui, Kai Dang, Xiaodong Deng, Yang Fan, Wenbin Ge, Yu~Han, Fei Huang, et~al. 2023.
\newblock Qwen technical report.
\newblock \emph{arXiv preprint arXiv:2309.16609}.

\bibitem[{Brown et~al.(2020)Brown, Mann, Ryder, Subbiah, Kaplan, Dhariwal, Neelakantan, Shyam, Sastry, Askell et~al.}]{brown2020language}
Tom Brown, Benjamin Mann, Nick Ryder, Melanie Subbiah, Jared~D Kaplan, Prafulla Dhariwal, Arvind Neelakantan, Pranav Shyam, Girish Sastry, Amanda Askell, et~al. 2020.
\newblock Language models are few-shot learners.
\newblock \emph{Advances in neural information processing systems}, 33:1877--1901.

\bibitem[{Chang et~al.(2023)Chang, Wang, Wang, Wu, Yang, Zhu, Chen, Yi, Wang, Wang, Ye, Zhang, Chang, Yu, Yang, and Xie}]{chang2023surveyevaluationlargelanguage}
Yupeng Chang, Xu~Wang, Jindong Wang, Yuan Wu, Linyi Yang, Kaijie Zhu, Hao Chen, Xiaoyuan Yi, Cunxiang Wang, Yidong Wang, Wei Ye, Yue Zhang, Yi~Chang, Philip~S. Yu, Qiang Yang, and Xing Xie. 2023.
\newblock \href {https://arxiv.org/abs/2307.03109} {A survey on evaluation of large language models}.
\newblock \emph{Preprint}, arXiv:2307.03109.

\bibitem[{Chen et~al.(2021)Chen, Tworek, Jun, Yuan, de~Oliveira~Pinto, Kaplan, Edwards, Burda, Joseph, Brockman, Ray, Puri, Krueger, Petrov, Khlaaf, Sastry, Mishkin, Chan, Gray, Ryder, Pavlov, Power, Kaiser, Bavarian, Winter, Tillet, Such, Cummings, Plappert, Chantzis, Barnes, Herbert-Voss, Guss, Nichol, Paino, Tezak, Tang, Babuschkin, Balaji, Jain, Saunders, Hesse, Carr, Leike, Achiam, Misra, Morikawa, Radford, Knight, Brundage, Murati, Mayer, Welinder, McGrew, Amodei, McCandlish, Sutskever, and Zaremba}]{chen2021codex}
Mark Chen, Jerry Tworek, Heewoo Jun, Qiming Yuan, Henrique~Ponde de~Oliveira~Pinto, Jared Kaplan, Harri Edwards, Yuri Burda, Nicholas Joseph, Greg Brockman, Alex Ray, Raul Puri, Gretchen Krueger, Michael Petrov, Heidy Khlaaf, Girish Sastry, Pamela Mishkin, Brooke Chan, Scott Gray, Nick Ryder, Mikhail Pavlov, Alethea Power, Lukasz Kaiser, Mohammad Bavarian, Clemens Winter, Philippe Tillet, Felipe~Petroski Such, Dave Cummings, Matthias Plappert, Fotios Chantzis, Elizabeth Barnes, Ariel Herbert-Voss, William~Hebgen Guss, Alex Nichol, Alex Paino, Nikolas Tezak, Jie Tang, Igor Babuschkin, Suchir Balaji, Shantanu Jain, William Saunders, Christopher Hesse, Andrew~N. Carr, Jan Leike, Josh Achiam, Vedant Misra, Evan Morikawa, Alec Radford, Matthew Knight, Miles Brundage, Mira Murati, Katie Mayer, Peter Welinder, Bob McGrew, Dario Amodei, Sam McCandlish, Ilya Sutskever, and Wojciech Zaremba. 2021.
\newblock \href {https://arxiv.org/abs/2107.03374} {Evaluating large language models trained on code}.

\bibitem[{Chen et~al.(2024)Chen, Lin, Sch{\"a}rli, and Zhou}]{chen2024teaching}
Xinyun Chen, Maxwell Lin, Nathanael Sch{\"a}rli, and Denny Zhou. 2024.
\newblock \href {https://openreview.net/forum?id=KuPixIqPiq} {Teaching large language models to self-debug}.
\newblock In \emph{The Twelfth International Conference on Learning Representations}.

\bibitem[{Clark et~al.(2019)Clark, Lee, Chang, Kwiatkowski, Collins, and Toutanova}]{clark-etal-2019-boolq}
Christopher Clark, Kenton Lee, Ming-Wei Chang, Tom Kwiatkowski, Michael Collins, and Kristina Toutanova. 2019.
\newblock \href {https://doi.org/10.18653/v1/N19-1300} {{B}ool{Q}: Exploring the surprising difficulty of natural yes/no questions}.
\newblock In \emph{Proceedings of the 2019 Conference of the North {A}merican Chapter of the Association for Computational Linguistics: Human Language Technologies, Volume 1 (Long and Short Papers)}, pages 2924--2936, Minneapolis, Minnesota. Association for Computational Linguistics.

\bibitem[{Cobbe et~al.(2021)Cobbe, Kosaraju, Bavarian, Chen, Jun, Kaiser, Plappert, Tworek, Hilton, Nakano, Hesse, and Schulman}]{cobbe2021gsm8k}
Karl Cobbe, Vineet Kosaraju, Mohammad Bavarian, Mark Chen, Heewoo Jun, Lukasz Kaiser, Matthias Plappert, Jerry Tworek, Jacob Hilton, Reiichiro Nakano, Christopher Hesse, and John Schulman. 2021.
\newblock Training verifiers to solve math word problems.
\newblock \emph{arXiv preprint arXiv:2110.14168}.

\bibitem[{DeepSeek-AI(2024)}]{deepseek-llm}
DeepSeek-AI. 2024.
\newblock \href {https://github.com/deepseek-ai/DeepSeek-LLM} {Deepseek llm: Scaling open-source language models with longtermism}.
\newblock \emph{arXiv preprint arXiv:2401.02954}.

\bibitem[{Dong et~al.(2024)Dong, Li, Dai, Zheng, Ma, Li, Xia, Xu, Wu, Chang, Sun, Li, and Sui}]{dong-etal-2024-survey}
Qingxiu Dong, Lei Li, Damai Dai, Ce~Zheng, Jingyuan Ma, Rui Li, Heming Xia, Jingjing Xu, Zhiyong Wu, Baobao Chang, Xu~Sun, Lei Li, and Zhifang Sui. 2024.
\newblock \href {https://doi.org/10.18653/v1/2024.emnlp-main.64} {A survey on in-context learning}.
\newblock In \emph{Proceedings of the 2024 Conference on Empirical Methods in Natural Language Processing}, pages 1107--1128, Miami, Florida, USA. Association for Computational Linguistics.

\bibitem[{Gao et~al.(2024{\natexlab{a}})Gao, Song, Miao, Cai, Yang, Chen, Hu, Xu, Dong, Zheng, Quan, Xiao, Zhang, Zan, Lu, Yu, Liu, Cui, Yang, Sha, Wang, Sui, Wang, Liu, and Chang}]{gao2024unifiedviewpreferencelearning}
Bofei Gao, Feifan Song, Yibo Miao, Zefan Cai, Zhe Yang, Liang Chen, Helan Hu, Runxin Xu, Qingxiu Dong, Ce~Zheng, Shanghaoran Quan, Wen Xiao, Ge~Zhang, Daoguang Zan, Keming Lu, Bowen Yu, Dayiheng Liu, Zeyu Cui, Jian Yang, Lei Sha, Houfeng Wang, Zhifang Sui, Peiyi Wang, Tianyu Liu, and Baobao Chang. 2024{\natexlab{a}}.
\newblock \href {https://arxiv.org/abs/2409.02795} {Towards a unified view of preference learning for large language models: A survey}.
\newblock \emph{Preprint}, arXiv:2409.02795.

\bibitem[{Gao et~al.(2024{\natexlab{b}})Gao, Song, Yang, Cai, Miao, Dong, Li, Ma, Chen, Xu et~al.}]{gao2024omni}
Bofei Gao, Feifan Song, Zhe Yang, Zefan Cai, Yibo Miao, Qingxiu Dong, Lei Li, Chenghao Ma, Liang Chen, Runxin Xu, et~al. 2024{\natexlab{b}}.
\newblock Omni-math: A universal olympiad level mathematic benchmark for large language models.
\newblock \emph{arXiv preprint arXiv:2410.07985}.

\bibitem[{GLM et~al.(2024)GLM, Zeng, Xu, Wang, Zhang, Yin, Zhang, Rojas, Feng, Zhao et~al.}]{glm2024chatglm}
Team GLM, Aohan Zeng, Bin Xu, Bowen Wang, Chenhui Zhang, Da~Yin, Dan Zhang, Diego Rojas, Guanyu Feng, Hanlin Zhao, et~al. 2024.
\newblock Chatglm: A family of large language models from glm-130b to glm-4 all tools.
\newblock \emph{arXiv preprint arXiv:2406.12793}.

\bibitem[{Han et~al.(2024)Han, Liang, Shi, He, and Xiao}]{han2024small}
Haixia Han, Jiaqing Liang, Jie Shi, Qianyu He, and Yanghua Xiao. 2024.
\newblock Small language model can self-correct.
\newblock In \emph{Proceedings of the AAAI Conference on Artificial Intelligence}, volume~38, pages 18162--18170.

\bibitem[{Hendrycks et~al.(2021)Hendrycks, Burns, Basart, Zou, Mazeika, Song, and Steinhardt}]{hendrycks2021measuring}
Dan Hendrycks, Collin Burns, Steven Basart, Andy Zou, Mantas Mazeika, Dawn Song, and Jacob Steinhardt. 2021.
\newblock \href {https://openreview.net/forum?id=d7KBjmI3GmQ} {Measuring massive multitask language understanding}.
\newblock In \emph{International Conference on Learning Representations}.

\bibitem[{Hu et~al.(2021)Hu, Shen, Wallis, Allen-Zhu, Li, Wang, Wang, and Chen}]{hu2021lora}
Edward~J Hu, Yelong Shen, Phillip Wallis, Zeyuan Allen-Zhu, Yuanzhi Li, Shean Wang, Lu~Wang, and Weizhu Chen. 2021.
\newblock Lora: Low-rank adaptation of large language models.
\newblock \emph{arXiv preprint arXiv:2106.09685}.

\bibitem[{Huang et~al.(2024)Huang, Chen, Mishra, Zheng, Yu, Song, and Zhou}]{huanglarge}
Jie Huang, Xinyun Chen, Swaroop Mishra, Huaixiu~Steven Zheng, Adams~Wei Yu, Xinying Song, and Denny Zhou. 2024.
\newblock \href {https://openreview.net/forum?id=IkmD3fKBPQ} {Large language models cannot self-correct reasoning yet}.
\newblock In \emph{The Twelfth International Conference on Learning Representations}.

\bibitem[{Ivison et~al.(2024)Ivison, Wang, Liu, Wu, Pyatkin, Lambert, Smith, Choi, and Hajishirzi}]{ivison2024unpacking}
Hamish Ivison, Yizhong Wang, Jiacheng Liu, Zeqiu Wu, Valentina Pyatkin, Nathan Lambert, Noah~A. Smith, Yejin Choi, and Hannaneh Hajishirzi. 2024.
\newblock \href {https://arxiv.org/abs/2406.09279} {Unpacking dpo and ppo: Disentangling best practices for learning from preference feedback}.
\newblock \emph{Preprint}, arXiv:2406.09279.

\bibitem[{Jiang et~al.(2023{\natexlab{a}})Jiang, Sablayrolles, Mensch, Bamford, Chaplot, Casas, Bressand, Lengyel, Lample, Saulnier et~al.}]{mistral}
Albert~Q Jiang, Alexandre Sablayrolles, Arthur Mensch, Chris Bamford, Devendra~Singh Chaplot, Diego de~las Casas, Florian Bressand, Gianna Lengyel, Guillaume Lample, Lucile Saulnier, et~al. 2023{\natexlab{a}}.
\newblock Mistral 7b.
\newblock \emph{arXiv preprint arXiv:2310.06825}.

\bibitem[{Jiang et~al.(2024)Jiang, Zhang, Weller, Weir, Van~Durme, and Khashabi}]{jiang2024self}
Dongwei Jiang, Jingyu Zhang, Orion Weller, Nathaniel Weir, Benjamin Van~Durme, and Daniel Khashabi. 2024.
\newblock Self-[in] correct: Llms struggle with refining self-generated responses.
\newblock \emph{CoRR}.

\bibitem[{Jiang et~al.(2023{\natexlab{b}})Jiang, Wang, and Wang}]{jiang2023selfevolve}
Shuyang Jiang, Yuhao Wang, and Yu~Wang. 2023{\natexlab{b}}.
\newblock Selfevolve: A code evolution framework via large language models.
\newblock \emph{arXiv preprint arXiv:2306.02907}.

\bibitem[{Kamoi et~al.(2024)Kamoi, Zhang, Zhang, Han, and Zhang}]{kamoi-etal-2024-llms}
Ryo Kamoi, Yusen Zhang, Nan Zhang, Jiawei Han, and Rui Zhang. 2024.
\newblock \href {https://doi.org/10.1162/tacl_a_00713} {When can {LLM}s actually correct their own mistakes? a critical survey of self-correction of {LLM}s}.
\newblock \emph{Transactions of the Association for Computational Linguistics}, 12:1417--1440.

\bibitem[{Kumar et~al.(2024)Kumar, Zhuang, Agarwal, Su, Co-Reyes, Singh, Baumli, Iqbal, Bishop, Roelofs et~al.}]{kumar2024training}
Aviral Kumar, Vincent Zhuang, Rishabh Agarwal, Yi~Su, John~D Co-Reyes, Avi Singh, Kate Baumli, Shariq Iqbal, Colton Bishop, Rebecca Roelofs, et~al. 2024.
\newblock Training language models to self-correct via reinforcement learning.
\newblock \emph{arXiv preprint arXiv:2409.12917}.

\bibitem[{Li et~al.(2024)Li, Chen, Chen, Zhang, Su, Xing, and Zhang}]{li2024confidence}
Loka Li, Zhenhao Chen, Guangyi Chen, Yixuan Zhang, Yusheng Su, Eric Xing, and Kun Zhang. 2024.
\newblock Confidence matters: Revisiting intrinsic self-correction capabilities of large language models.
\newblock \emph{arXiv preprint arXiv:2402.12563}.

\bibitem[{Liu et~al.(2024)Liu, Nassereldine, Yang, Xu, Hu, Li, Kumar, Lee, and Xiong}]{liu2024large}
Dancheng Liu, Amir Nassereldine, Ziming Yang, Chenhui Xu, Yuting Hu, Jiajie Li, Utkarsh Kumar, Changjae Lee, and Jinjun Xiong. 2024.
\newblock Large language models have intrinsic self-correction ability.
\newblock \emph{CoRR}.

\bibitem[{Madaan et~al.(2024)Madaan, Tandon, Gupta, Hallinan, Gao, Wiegreffe, Alon, Dziri, Prabhumoye, Yang et~al.}]{madaan2024self}
Aman Madaan, Niket Tandon, Prakhar Gupta, Skyler Hallinan, Luyu Gao, Sarah Wiegreffe, Uri Alon, Nouha Dziri, Shrimai Prabhumoye, Yiming Yang, et~al. 2024.
\newblock Self-refine: Iterative refinement with self-feedback.
\newblock \emph{Advances in Neural Information Processing Systems}, 36.

\bibitem[{Ouyang et~al.(2022)Ouyang, Wu, Jiang, Almeida, Wainwright, Mishkin, Zhang, Agarwal, Slama, Ray, Schulman, Hilton, Kelton, Miller, Simens, Askell, Welinder, Christiano, Leike, and Lowe}]{ouyang2022traininglanguagemodelsfollow}
Long Ouyang, Jeff Wu, Xu~Jiang, Diogo Almeida, Carroll~L. Wainwright, Pamela Mishkin, Chong Zhang, Sandhini Agarwal, Katarina Slama, Alex Ray, John Schulman, Jacob Hilton, Fraser Kelton, Luke Miller, Maddie Simens, Amanda Askell, Peter Welinder, Paul Christiano, Jan Leike, and Ryan Lowe. 2022.
\newblock \href {https://arxiv.org/abs/2203.02155} {Training language models to follow instructions with human feedback}.
\newblock \emph{Preprint}, arXiv:2203.02155.

\bibitem[{Pan and Zeng(2023)}]{pan2023llmspossesspersonalitymaking}
Keyu Pan and Yawen Zeng. 2023.
\newblock \href {https://arxiv.org/abs/2307.16180} {Do llms possess a personality? making the mbti test an amazing evaluation for large language models}.
\newblock \emph{Preprint}, arXiv:2307.16180.

\bibitem[{Pan et~al.(2024)Pan, Saxon, Xu, Nathani, Wang, and Wang}]{pan-etal-2024-automatically}
Liangming Pan, Michael Saxon, Wenda Xu, Deepak Nathani, Xinyi Wang, and William~Yang Wang. 2024.
\newblock \href {https://doi.org/10.1162/tacl_a_00660} {Automatically correcting large language models: Surveying the landscape of diverse automated correction strategies}.
\newblock \emph{Transactions of the Association for Computational Linguistics}, 12:484--506.

\bibitem[{Qu et~al.(2024)Qu, Zhang, Garg, and Kumar}]{qurecursive}
Yuxiao Qu, Tianjun Zhang, Naman Garg, and Aviral Kumar. 2024.
\newblock \href {https://openreview.net/forum?id=qDXdmdBLhR} {Recursive introspection: Teaching foundation model agents how to self-improve}.
\newblock In \emph{Automated Reinforcement Learning: Exploring Meta-Learning, AutoML, and LLMs}.

\bibitem[{Radford et~al.(2018)Radford, Narasimhan, Salimans, Sutskever et~al.}]{radford2018improving}
Alec Radford, Karthik Narasimhan, Tim Salimans, Ilya Sutskever, et~al. 2018.
\newblock Improving language understanding by generative pre-training.

\bibitem[{Radford et~al.(2019)Radford, Wu, Child, Luan, Amodei, Sutskever et~al.}]{radford2019language}
Alec Radford, Jeffrey Wu, Rewon Child, David Luan, Dario Amodei, Ilya Sutskever, et~al. 2019.
\newblock Language models are unsupervised multitask learners.
\newblock \emph{OpenAI blog}, 1(8):9.

\bibitem[{Rafailov et~al.(2024)Rafailov, Sharma, Mitchell, Ermon, Manning, and Finn}]{rafailov2024directpreferenceoptimizationlanguage}
Rafael Rafailov, Archit Sharma, Eric Mitchell, Stefano Ermon, Christopher~D. Manning, and Chelsea Finn. 2024.
\newblock \href {https://arxiv.org/abs/2305.18290} {Direct preference optimization: Your language model is secretly a reward model}.
\newblock \emph{Preprint}, arXiv:2305.18290.

\bibitem[{Stechly et~al.(2023)Stechly, Marquez, and Kambhampati}]{stechly2023gpt}
Kaya Stechly, Matthew Marquez, and Subbarao Kambhampati. 2023.
\newblock Gpt-4 doesn’t know it’s wrong: An analysis of iterative prompting for reasoning problems.
\newblock In \emph{NeurIPS 2023 Foundation Models for Decision Making Workshop}.

\bibitem[{Talmor et~al.(2019)Talmor, Herzig, Lourie, and Berant}]{talmor-etal-2019-commonsenseqa}
Alon Talmor, Jonathan Herzig, Nicholas Lourie, and Jonathan Berant. 2019.
\newblock \href {https://doi.org/10.18653/v1/N19-1421} {{C}ommonsense{QA}: A question answering challenge targeting commonsense knowledge}.
\newblock In \emph{Proceedings of the 2019 Conference of the North {A}merican Chapter of the Association for Computational Linguistics: Human Language Technologies, Volume 1 (Long and Short Papers)}, pages 4149--4158, Minneapolis, Minnesota. Association for Computational Linguistics.

\bibitem[{Tyen et~al.(2024)Tyen, Mansoor, Carbune, Chen, and Mak}]{tyen-etal-2024-llms}
Gladys Tyen, Hassan Mansoor, Victor Carbune, Peter Chen, and Tony Mak. 2024.
\newblock \href {https://doi.org/10.18653/v1/2024.findings-acl.826} {{LLM}s cannot find reasoning errors, but can correct them given the error location}.
\newblock In \emph{Findings of the Association for Computational Linguistics: ACL 2024}, pages 13894--13908, Bangkok, Thailand. Association for Computational Linguistics.

\bibitem[{Valmeekam et~al.(2023)Valmeekam, Marquez, and Kambhampati}]{valmeekam2023can}
Karthik Valmeekam, Matthew Marquez, and Subbarao Kambhampati. 2023.
\newblock Can large language models really improve by self-critiquing their own plans?
\newblock In \emph{NeurIPS 2023 Foundation Models for Decision Making Workshop}.

\bibitem[{Wei et~al.(2021)Wei, Bosma, Zhao, Guu, Yu, Lester, Du, Dai, and Le}]{weifinetuned}
Jason Wei, Maarten Bosma, Vincent Zhao, Kelvin Guu, Adams~Wei Yu, Brian Lester, Nan Du, Andrew~M Dai, and Quoc~V Le. 2021.
\newblock Finetuned language models are zero-shot learners.
\newblock In \emph{International Conference on Learning Representations}.

\bibitem[{Welleck et~al.(2023)Welleck, Lu, West, Brahman, Shen, Khashabi, and Choi}]{welleck2023generating}
Sean Welleck, Ximing Lu, Peter West, Faeze Brahman, Tianxiao Shen, Daniel Khashabi, and Yejin Choi. 2023.
\newblock \href {https://openreview.net/forum?id=hH36JeQZDaO} {Generating sequences by learning to self-correct}.
\newblock In \emph{The Eleventh International Conference on Learning Representations}.

\bibitem[{Wu et~al.(2024)Wu, Zeng, Zhang, Tan, Shen, and Jiang}]{wu-etal-2024-large}
Zhenyu Wu, Qingkai Zeng, Zhihan Zhang, Zhaoxuan Tan, Chao Shen, and Meng Jiang. 2024.
\newblock \href {https://doi.org/10.18653/v1/2024.emnlp-main.714} {Large language models can self-correct with key condition verification}.
\newblock In \emph{Proceedings of the 2024 Conference on Empirical Methods in Natural Language Processing}, pages 12846--12867, Miami, Florida, USA. Association for Computational Linguistics.

\bibitem[{Xi et~al.(2023)Xi, Jin, Zhou, Zheng, Gao, Liu, Gui, Zhang, and Huang}]{xi-etal-2023-self}
Zhiheng Xi, Senjie Jin, Yuhao Zhou, Rui Zheng, Songyang Gao, Jia Liu, Tao Gui, Qi~Zhang, and Xuanjing Huang. 2023.
\newblock \href {https://doi.org/10.18653/v1/2023.findings-emnlp.762} {Self-{P}olish: Enhance reasoning in large language models via problem refinement}.
\newblock In \emph{Findings of the Association for Computational Linguistics: EMNLP 2023}, pages 11383--11406, Singapore. Association for Computational Linguistics.

\bibitem[{Yan et~al.(2024)Yan, Jiang, Liu, Cao, Xu, Cai, Shao et~al.}]{yan2024s}
Yuchen Yan, Jin Jiang, Yang Liu, Yixin Cao, Xin Xu, Xunliang Cai, Jian Shao, et~al. 2024.
\newblock Sc-math: Spontaneous step-level self-correction makes large language models better mathematical reasoners.
\newblock \emph{arXiv preprint arXiv:2409.01524}.

\bibitem[{Yang et~al.(2024{\natexlab{a}})Yang, Yang, Hui, Zheng, Yu, Zhou, Li, Li, Liu, Huang et~al.}]{yang2024qwen2}
An~Yang, Baosong Yang, Binyuan Hui, Bo~Zheng, Bowen Yu, Chang Zhou, Chengpeng Li, Chengyuan Li, Dayiheng Liu, Fei Huang, et~al. 2024{\natexlab{a}}.
\newblock Qwen2 technical report.
\newblock \emph{arXiv preprint arXiv:2407.10671}.

\bibitem[{Yang et~al.(2023)Yang, Dai, Wang, and Sui}]{yang-etal-2023-demonstration}
Zhe Yang, Damai Dai, Peiyi Wang, and Zhifang Sui. 2023.
\newblock \href {https://doi.org/10.18653/v1/2023.findings-emnlp.880} {Not all demonstration examples are equally beneficial: Reweighting demonstration examples for in-context learning}.
\newblock In \emph{Findings of the Association for Computational Linguistics: EMNLP 2023}, pages 13209--13221, Singapore. Association for Computational Linguistics.

\bibitem[{Yang et~al.(2024{\natexlab{b}})Yang, Zhang, Liu, Yang, Lin, Zhou, and Sui}]{yang-etal-2024-large-language-models-always}
Zhe Yang, Yichang Zhang, Tianyu Liu, Jian Yang, Junyang Lin, Chang Zhou, and Zhifang Sui. 2024{\natexlab{b}}.
\newblock \href {https://doi.org/10.18653/v1/2024.emnlp-main.92} {Can large language models always solve easy problems if they can solve harder ones?}
\newblock In \emph{Proceedings of the 2024 Conference on Empirical Methods in Natural Language Processing}, pages 1531--1555, Miami, Florida, USA. Association for Computational Linguistics.

\bibitem[{Zhang et~al.(2023)Zhang, Dong, Li, Zhang, Sun, Wang, Li, Hu, Zhang, Wu et~al.}]{zhang2023instruction}
Shengyu Zhang, Linfeng Dong, Xiaoya Li, Sen Zhang, Xiaofei Sun, Shuhe Wang, Jiwei Li, Runyi Hu, Tianwei Zhang, Fei Wu, et~al. 2023.
\newblock Instruction tuning for large language models: A survey.
\newblock \emph{arXiv preprint arXiv:2308.10792}.

\bibitem[{Zhang et~al.(2024{\natexlab{a}})Zhang, Shen, Wu, Peng, Wang, Zhuang, and Lu}]{zhang-etal-2024-self-contrast}
Wenqi Zhang, Yongliang Shen, Linjuan Wu, Qiuying Peng, Jun Wang, Yueting Zhuang, and Weiming Lu. 2024{\natexlab{a}}.
\newblock \href {https://doi.org/10.18653/v1/2024.acl-long.197} {Self-contrast: Better reflection through inconsistent solving perspectives}.
\newblock In \emph{Proceedings of the 62nd Annual Meeting of the Association for Computational Linguistics (Volume 1: Long Papers)}, pages 3602--3622, Bangkok, Thailand. Association for Computational Linguistics.

\bibitem[{Zhang et~al.(2024{\natexlab{b}})Zhang, Khalifa, Logeswaran, Kim, Lee, Lee, and Wang}]{zhang-etal-2024-small}
Yunxiang Zhang, Muhammad Khalifa, Lajanugen Logeswaran, Jaekyeom Kim, Moontae Lee, Honglak Lee, and Lu~Wang. 2024{\natexlab{b}}.
\newblock \href {https://doi.org/10.18653/v1/2024.findings-acl.924} {Small language models need strong verifiers to self-correct reasoning}.
\newblock In \emph{Findings of the Association for Computational Linguistics: ACL 2024}, pages 15637--15653, Bangkok, Thailand. Association for Computational Linguistics.

\bibitem[{Zhou et~al.(2023)Zhou, Lu, Mishra, Brahma, Basu, Luan, Zhou, and Hou}]{zhou2023instruction}
Jeffrey Zhou, Tianjian Lu, Swaroop Mishra, Siddhartha Brahma, Sujoy Basu, Yi~Luan, Denny Zhou, and Le~Hou. 2023.
\newblock Instruction-following evaluation for large language models.
\newblock \emph{arXiv preprint arXiv:2311.07911}.

\end{thebibliography}
\clearpage
\appendix
\section*{Appendix}
\label{sec:appendix}

\section{Mathematical Notations}
\label{app:notations}
This section shows all of the mathematical notations used in this paper. If you forget the meaning of any notation, please refer to Table \ref{tab:notations}. We leverage $\ \hat{} \  $ to symbolize estimates (e.g. $\hat{P}(a_i)$ represents the estimate of the true value $P(a_i)$ ). For simplicity, we only show true values in Table \ref{tab:notations}, and estimates are omitted. 

\begin{table*}[!tb]
\centering
\footnotesize
\resizebox{\textwidth}{!}{
\begin{tabular}{l|p{0.65\linewidth}}
\toprule
\textbf{Notations} & \textbf{Meanings} \\
\midrule
$A$ & question set \\
\cmidrule{1-2}
$q_i$ & the $i^{th}$ question in $A$\\
\cmidrule{1-2}
$P(a_i)$ & the probability of generating a correct answer for question $q_i$ through a single temperature-based sampling before self-correction \\
\cmidrule{1-2}
$P(b_i)$ & the probability of generating a correct answer for question $q_i$ through a single temperature-based sampling after self-correction \\
\cmidrule{1-2}
$P(a)$ & the probability of generating a correct answer for a random question $q$ in $A$ through a single temperature-based sampling before self-correction \\
\cmidrule{1-2}
$P(b)$ & the probability of generating a correct answer for a random question $q$ in $A$ through a single temperature-based sampling after self-correction \\
\cmidrule{1-2}
$P(b_i|a_i)$ & the conditional probability of generating a correct answer after self-correction, given the initial answer is correct \\
\cmidrule{1-2}
$P(b_i|\neg a_i)$ & the conditional probability of generating a correct answer after self-correction, given the initial answer is incorrect \\
\cmidrule{1-2}
$Acc_1$ & accuracy before self-correction (i.e. expectation of $P(a)$) \\
\cmidrule{1-2}
$Acc_2$ & accuracy before self-correction (i.e. expectation of $P(b)$) \\
\cmidrule{1-2}
$Acc_2^{low}$ & lower bound of $Acc_2$ \\
\cmidrule{1-2}
$Acc_2^{upp}$ & upper bound of $Acc_2$ \\
\bottomrule
\end{tabular}
}
\caption{Mathematical notations and their meanings.}
\label{tab:notations}
\end{table*}

\section{Metric Derivation Details}
\label{app:metric_derivation}
This section shows a detailed derivation of Confidence Score (\S \ref{subapp:CL_derivation}) and Critique Score (\S \ref{subapp:CS_derivation}), along with the proof of Equation \ref{equ:weighted_sum} (\S \ref{subapp:weighted_sum_proof}).

\subsection{Derivation of CL}
\label{subapp:CL_derivation}
Let's think about the stochastic process defined in \S \ref{subsec:notations}: 

\textbullet~ Randomly sampling a question $q$ from $A$ with equal probability.

Initially, the prior probability of selecting $q_i$ in the above random process is $P(select \  q_i) = \frac{1}{n}$. After introducing the condition that the model has answered question $q_i$ correctly initially, the posterior probability of $q_i$ being selected in the random process becomes $P(select \  q_i) = \frac{P(a_i)}{\sum_{j=1,...,n}P(a_j)}$. By leveraging this posterior probability for the calculation of expected values, we have:

\begin{equation}
\label{equ:CL_derivation}
\begin{aligned}
CL &= E[P(b|a)] \\
&= \sum_{i=1,...N}P(select \  q_i)P(b_i|a_i) \\
&= \sum_{i=1,...,n}\frac{P(a_i)}{\sum_{j=1,...,n}P(a_j)}P(b_i|a_i) \\
&= \frac{\sum_{i=1,...,n}P(a_i)P(b_i|a_i)} {\sum_{i=1,...,n}P(a_i)},
\end{aligned}
\end{equation}

where $P(b_i|a_i)$ is the conditional probability of a model correctly answering $q_i$ after self-correction given that it has correctly answered it initially. The higher CL is, the more confident the model is about its correct answers. High CL also indicates the model is confident and will not change its correct answer even when challenged.

\subsection{Derivation of CS}
\label{subapp:CS_derivation}
We can derive CS in a manner similar to Equation \ref{equ:CL_derivation}, but here we would give another form of derivation:

\begin{equation}
\begin{aligned}
CC &= E[P(b|\neg a)] \\
&= E[\frac{P(b,\neg a)}{P(\neg a)}] \\
&= \frac{\sum_{i=1,...,n}P(b_i,\neg a_i)/N}{\sum_{i=1,...,n}P(\neg a_i)/N} \\
&= \frac{\sum_{i=1,...,n}P(b_i|\neg a_i)P(\neg a_i)}{\sum_{i=1,...,n}P(\neg a_i)} \\
&= \frac{\sum_{i=1,...,n}[1-P(a_i)]P(b_i|\neg a_i)}{\sum_{i=1,...,n}[1-P(a_i)]},
\end{aligned}
\end{equation}

where $P(b_i|\neg a_i)$ is the conditional probability of a model correctly answering $a_i$ after self-correction given that it has answered it wrong initially, and model answer $a_i$ wrong with probability $P(\neg a_i) = 1 - P(a_i)$. 
CS reflects the extent to which the model persists in providing wrong answers. A lower CS value indicates a greater tendency for the model to stubbornly maintain erroneous responses, whereas a higher CS value suggests a greater willingness of the model to correct these errors.
\subsection{Proof of Equation \ref{equ:weighted_sum}}
\label{subapp:weighted_sum_proof}
How can we ensure that a model maintains a high accuracy after self-correction? According to the probability decomposition formula, we have:
$$P(b_i)=P(b_i|a_i)P(a_i)+P(b_i|\neg a_i)P(\neg a_i),$$ 
which indicates: 
(1) In the scenario where the model provides a correct answer initially,
high confidence in its answer will lead to a low likelihood of changing its response, and consequently results in a high probability of correctness after self-correction;
(2) Conversely, if the model initially provides an incorrect answer, it has the opportunity to correct its error after self-correction, which also facilitates a higher likelihood of giving a correct answer.

Based on these observations, it can be intuitively concluded that higher values of $CL$ and $CS$ will lead to an increase in $Acc_2$. Besides, we also discover the following mathematical relationships:

\begin{equation}
\label{equ:weighted_sum_derivation}
\begin{aligned}
& Acc_2 \\
&= \frac{\sum_{i=1}^{n}P(b_i)}{n} \\ 
&= \frac{\sum_{i=1}^{n}P(b_i|a_i)P(a_i)+P(b_i|\neg a_i)P(\neg a_i)}{n} \\
&= \frac{\sum_{i=1}^{n}P(a_i)}{n}\frac{\sum_{i=1}^{n}P(a_i)P(b_i|a_i)}{\sum_{i=1}^{n}P(a_i)} \\
& + \frac{\sum_{i=1}^{n}[1-P(a_i)]}{n}\frac{\sum_{i=1}^{n}P(\neg a_i)P(b_i|\neg a_i)}{\sum_{i=1}^{n}[1-P(a_i)]} \\
&=Acc_1*CL+(1-Acc_1)*CS
\end{aligned}
\end{equation}

\section{Derivation of RSS}
\label{app:RSS_derivation}


The derivation of Relative Self-correction Score (RSS) can be summarized as follows: Initially, we utilize an assumed inequation to estimate the possible range of of CL and CS. Subsequently, by using Equation \ref{equ:weighted_sum}, we determine the corresponding range for $Acc_2$, thus obtaining the upper and lower bounds for $Acc_2$, and ultimately deriving the final RSS.

From a probabilistic perspective, $Acc_1$, CL, and CS are interpreted as follows: $Acc_1$ represents the probability that the model correctly answers a question without any conditions. In contrast, CL and CS represent the conditional probabilities that the model correctly answer the question given that it has previously answered it right or wrong, respectively. For questions the model is already capable of answering correctly, there is a higher likelihood of continuing to do so. Conversely, for questions the model initially answers incorrectly, the probability of subsequently correcting is lower.
Based on the this analysis, we assume the following inequality holds:
\[
CS \leq Acc_1 \leq CL
\]

Experimental results in \S \ref{sec:experiments} also empirically demonstrate that this inequality is valid. So we have $CS \in [0,Acc_1]$ and $CL \in [Acc_1,1]$. By substituting $CS=0$ and $CL=Acc_1$ into Equation \ref{equ:weighted_sum}, we have the lower bound for $Acc_2$ is:
\[
Acc_2^{\text{low}} = Acc_1 \cdot Acc_1 + (1-Acc_1) \cdot 0 = Acc_1^2
\]

By substituting $CS=Acc_1$ and $CL=1$ into Equation \ref{equ:weighted_sum}, the upper bound for $Acc_2$ becomes:
\[
Acc_2^{\text{upp}} = Acc_1 \cdot 1 + (1-Acc_1) \cdot Acc_1 = 2Acc_1 - Acc_1^2
\]

We define RSS as the normalized $Acc_2$, indicating its position within the aforementioned interval:
\[
RSS = \frac{Acc_2 - Acc_2^{\text{low}}}{Acc_2^{\text{upp}} - Acc_2^{\text{low}}} = \frac{Acc_2 - Acc_1^2}{2Acc_1-2Acc_1^2}
\]

\section{Probability Estimation}
\label{app:probability_estimation}



Metrics in \S \ref{sec:methodology} are derived from a probabilistic perspective, and their calculation relies on 3 key probability values \( P(a_i), P(b_i|a_i) \) and \( P(b_i| \neg a_i) \) of each question $q_i$. However, the actual values of these probabilities are unattainable. In practice, we utilize statistical methods to obtain their estimates \( \hat{P}(a), \hat{P}(b_i|a_i) \) and \( \hat{P}(b_i| \neg a_i) \) to substitute these true values for metric computation.
Currently, natural language processing (NLP) tasks can be generally divided into classification tasks and generation tasks, and we will discuss the probability estimation methods applied to these two types of tasks separately.

\paragraph{Probability Estimation for Classification Tasks.} For a \( K \)-class classification task, let the set of all candidate labels be denoted as \( L = \{l_0, l_1, \ldots, l_{K-1}\} \) (e.g., for the MMLU, the candidate set is \( \{A, B, C, D\} \)). A question $q_i$ is fed into the model and the model is asked to output the predicted label. When the model performs next-token prediction, it first generates a logit vector $(o_0,o_1,...o_{|V|-1})$, where each value corresponds to the logit of a token in the vocabulary \(V\) and \(|V|\) denotes the size of the vocabulary. In the generation process, The logit vector is then passed through a softmax layer to produce the probability distribution of the next token in the whole vocabulary.
However, for classification tasks, we are only interested in the probability distribution over candidate label set $L$ instead of vocabulary $V$. Therefore, we discard most logit values, retaining only those corresponding to candidate labels, resulting in a reduced logit vector $(o^{'}_0, o^{'}_1, \ldots, o^{'}_{K-1})$. After applying the softmax layer, the model predicts the probability for each label \( P(l_0), P(l_1), \ldots, P(l_{K-1}) \). 

(1). Without any loss of generality, assume the correct label is \( l_0 \), then we have \( \hat{P}(a_i) = P(l_0) \).

(2). Next, we feed the correct answer \( l_0 \) into the model and ask the model to self-correct. The model outputs the probability distribution over candidate labels, denoted as \( P(l_0|l_0), P(l_1|l_0), \ldots, P(l_{K-1}|l_0) \), then we have \( \hat{P}(b_i|a_i) = P(l_0|l_0) \).

(3). The computation of \( \hat{P}(b_i|\neg a_i) \) is more complex. For each incorrect label \( l_j \) (\( j \neq 0 \)), we input it to the model and allow for self-correction, obtaining the probability of correcting it to the correct label \( P(l_0|l_j) \). Finally, by using the law of total probability, we have \( \hat{P}(b_i|\neg a_i) = \sum_{j=1,\ldots,K-1} P(l_0|l_j)P(l_j) \).

\paragraph{Probability Estimation for Generation Task.} We employ multiple sampling to estimate probabilities by observing the frequency of correct and incorrect answers. Given a question \(q_i\), we pose it to the model and obtain an initial answer. Subsequently, the model is prompted to self-correct the initial answer, resulting in a final answer. This process is repeated \(T\) times, and for each pair of initial and final answers, we evaluate their correctness. This yields a sequence of results \((a_i^0, b_i^0), (a_i^1, b_i^1), \ldots, (a_i^{T-1}, b_i^{T-1})\), where \((a_i^t, b_i^t)\) denotes the outcome of the \(t\)-th repetition. Specifically, \(a_i^t\) and \(b_i^t\) indicate the correctness of the initial and final answers, respectively. For a correct initial answer, \(a_i^t = 1\); otherwise, \(a_i^t = 0\) and The same logic applies to \(b_i^t\). Utilizing frequency to estimate probability, we have: 

(1). \(\hat{P}(a_i) =  \frac{\sum_{t=0}^{T-1}a_i^t}{T}\);

(2). \(\hat{P}(b_i|a_i) = \frac{\sum_{t=0}^{T-1} a_i^t b_i^t}{\sum_{t=0}^{T-1} a_i^t}\);

(3). \(\hat{P}(b_i|\neg a_i) = \frac{\sum_{t=0}^{T-1} (1-a_i^t)b_i^t}{\sum_{t=0}^{T-1} (1-a_i^t)}\)

\section{Metric Convergence}
\label{app:convergence}
\begin{figure}[!htb]
    \centering
    \includegraphics[width=0.48\textwidth]{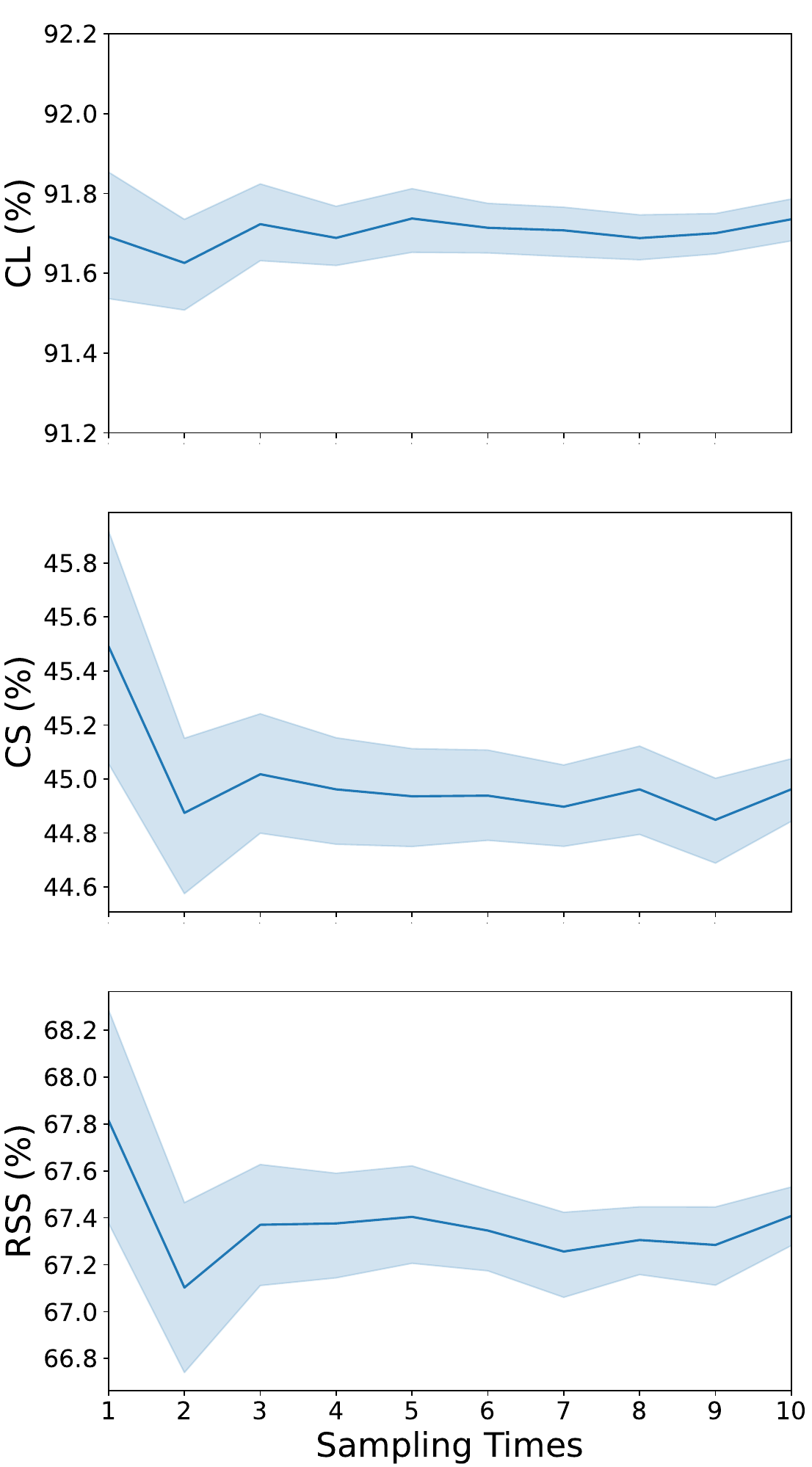}
    \caption{Metric convergence for sampling-based probability estimation method.}
    \label{fig:convergence}
\end{figure}

We study the convergence of our proposed three metrics for sampling-based probability estimation method. Taking experimental results for Llama3-8B-Instruct on GSM8k shown in Figure \ref{fig:convergence} as an example,
our metrics can converge and arrive at relatively stable values through about 3 times sampling.
\section{Implementation Details}

\subsection{More Implementation Details for \S \ref{sec:experiments}}
\label{subapp:setup1}
Most of these open-source models are released with two versions, the pre-trained base model and the chat model (base model + instruction tuning and alignment), and we focus our evaluation solely on chat models
For classification tasks, we estimate probability by logits; for generation tasks, we estimate probability by multiple samplings, and more details about probability estimation methods are available in Appendix \ref{app:probability_estimation}. For each question, we repeatedly sample 10 times with default sampling hyper-parameters (e.g. temperature) released by model developers. For each small open-source model (< 10B), we run the experiments on a single Nvidia A100 80G GPU; for each large model (about 70B), experiments are conducted on 4 Nvidia A100 80G GPUs. For faster generation speed, we utilize vllm \footnote{\url{https://github.com/vllm-project/vllm}} to accelerate.

For closed-source models whose logits are unavailable, we treat classification tasks as generation tasks and estimate probability by sampling. To reduce API calls, we only sample 3 times for each question. For a dataset with more than 500 items, we randomly sample 500 items and test on this subset. There are also different versions of closed-source models, and we utilize the latest version of GPT-3.5 Turbo (gpt-3.5-turbo-0125) and GPT-4 Turbo (gpt-4-turbo-2024-04-09).


\subsection{More Implementation Details for \S \ref{sec:tuning}}
\label{subapp:setup2}
For GSM8k, we sample multiple answers for each question by Llama3-8B-Instruct to build an answer base, then select correct-correct answer pairs to construct CLT data and correct-wrong answer pairs to construct CST data, which is similar to \citet{welleck2023generating,kumar2024training}.
For MMLU and BoolQ, we construct CLT and CST automatically from the original training data (choosing the correct answer twice for CLT and choosing the correct and a random wrong answer from candidates).

We train models through the implementation provided by \citet{ivison2024unpacking} \footnote{\url{https://github.com/allenai/open-instruct}}. For BoolQ and GSM8k, we train 2 epochs; for MMLU we train only 1 epoch due to the large training set. More training hyper-parameters are shown in Table \ref{tab:training_hyper}.
\begin{table}[!ht]
  \centering
  \footnotesize
    \begin{tabular}{l|c}
    \toprule
    learning rate & 5e-5 \\
    lr scheduler & cosine \\
    mixed precision & bf16 \\
    weight decay & 0.0 \\
    warmup ratio & 0.0 \\
    lora rank & 64 \\
    lora alpha & 16 \\
    lora dropout & 0.1 \\ 
    \bottomrule
    \end{tabular}
\caption{Training hyper-parameters.}
\label{tab:training_hyper}
\end{table}

\section{More Experimental Results}
\label{app:experiment_results}
We show more experimental results in this section: Experiment results on IFEval, Humaneval, and CommonsenseQA are shown in Table \ref{tab:main_results_more}; relative self-correction score results are shown in Table \ref{tab:RSS_results}.

\begin{table*}[htbp]
  \centering
  \footnotesize
    \begin{tabular}{l|cccc|cccc|cccc}
    \toprule
    \multicolumn{1}{c|}{\multirow{2}[2]{*}{Models}} & \multicolumn{4}{c|}{IFEval}   & \multicolumn{4}{c|}{Humaneval} & \multicolumn{4}{c}{CommonsenseQA} \\
    & $Acc_1$  & $Acc_2$  & $CL$    & $CS$    & $Acc_1$  & $Acc_2$  & $CL$    & $CS$    & $Acc_1$  & $Acc_2$  & $CL$   & $CS$  \\
    \midrule
    Llama3-8B-Instruct & 64.0  & 70.1  & 92.8  & 29.7  & 52.7  & 50.1  & 77.7  & 19.4  & 74.7  & 76.7  & 94.9  & 23.0  \\
    Deepseek-7B-Chat & 37.4  & 38.6  & 93.0  & 6.1   & 39.7  & 39.9  & 99.7  & 0.6   & 67.1  & 67.4  & 99.7  & 1.3  \\
    Mistral-7B-Instruct & 44.2  & 43.6  & 90.7  & 6.3   & 32.4  & 32.1  & 84.8  & 6.8   & 70.0  & 71.2  & 99.0  & 6.5  \\
    Qwen2.5-7B-Chat & 71.7  & 74.8  & 96.1  & 20.8  & 74.3  & 75.3  & 96.5  & 14.0  & 82.6  & 82.0  & 93.6  & 26.9  \\
    GLM4-9B-Chat & 29.9  & 31.0  & 90.5  & 5.6   & 64.9  & 63.7  & 86.9  & 20.7  & 77.8  & 78.8  & 87.0  & 50.0  \\
    \midrule
    Llama3-70B-Instruct & 76.0  & 80.5  & 96.4  & 30.1  & 74.8  & 69.9  & 84.8  & 25.8  & 82.1  & 83.7  & 97.1  & 22.3  \\
    Deepseek-67B-Chat & 51.0  & 51.9  & 96.7  & 5.3   & 65.2  & 65.0  & 97.2  & 4.7   & 74.4  & 76.2  & 95.4  & 20.5  \\
    Qwen2.5-72B-Chat & 84.7  & 84.8  & 97.1  & 17.3  & 81.7  & 81.3  & 97.5  & 8.9   & 85.5  & 86.7  & 98.4  & 18.0  \\
    \midrule
    Qwen-Max & 83.4  & 85.2  & 97.9  & 21.6  & 80.9  & 81.5  & 96.2  & 19.1  &  90.1   & 88.5  & 97.0  & 10.7  \\
    GPT-3.5 Turbo  & 65.9  & 67.7  & 94.2  & 16.6  & 64.4  & 66.3  & 91.5  & 20.6  &  79.9   & 76.2   & 86.7   & 34.4  \\
    GPT-4 Turbo  & 79.1  & 81.9  & 96.7  & 26.2  & 82.5  & 83.9  & 95.8  & 27.9  &   85.0   & 77.4  & 81.7  & 52.9 \\
    \bottomrule
    \end{tabular}%
\caption{Experiment results on IFEval, Humaneval and CommonsenseQA. We report accuracy(\%) before and after self-correction (denoted as $Acc_1$ and $Acc_2$). Confidence Level ($CL$) and Critique Score ($CS$) are also shown for fine-grained analysis of self-correction behavior.}
  \label{tab:main_results_more}%
\end{table*}%

\begin{table*}[!htb]
  \centering
  \footnotesize
    \begin{tabular}{l|cccccc}
    \toprule
    \multicolumn{1}{c|}{Models} & GSM8k & IFEval & Humaneval & MMLU  & BoolQ & CommensenseQA \\
    \midrule
    Llama3-8B-Instruct & 67.3  & 63.3  & 44.7  & 53.9  & 55.3  & 55.3 \\
    Deepseek-7B-Chat & 49.3  & 52.5  & 50.5  & 50.1  & 49.5  & 50.5 \\
    Mistral-7B-Instruct-v3 & 51.9  & 50.0    & 49.2  & 50.0    & 52.4  & 52.9 \\
    Qwen2.5-7B-Chat & 54.1  & 57.7  & 52.6  & 51.2  & 54.3  & 47.9 \\
    GLM4-9B-Chat & 56.7  & 52.6  & 47.3  & 52.4  & 57.7  & 52.7 \\
    \midrule
    Llama3-70B-Instruct & 61.8  & 62.3  & 37.1  & 53.8  & 50.1  & 55.5 \\
    Deepseek-67B-Chat & 49.7  & 51.8  & 49.6  & 52.3  & 50.0    & 54.7 \\
    Qwen2.5-72B-Chat & 52.5  & 50.7  & 48.6  & 52.9  & 72.8  & 55.1 \\
    Qwen-Max & 55.0    & 56.5  & 52.0   & 54.5  & 55.7  &  41.1 \\
    GPT-3.5 Turbo  & 59.0    & 54.1  & 54.0    & 50.7  & 54.3  & 38.6 \\
    GPT-4 Turbo  & 38.1  & 58.7  & 54.9  & 42.7  & 44.1  & 20.2 \\
    \bottomrule
    \end{tabular}%
  \caption{Relative Self-correction Score results.}
  \label{tab:RSS_results}%
\end{table*}%


\section{Prompt}
\label{app:prompt}
We show the prompts utilized in \S \ref{sec:Analysis} for LLM self-correction behavior manipulation in Table \ref{tab:prompt}.

\begin{table*}[!htb]
\centering
\resizebox{\textwidth}{!}{
\begin{tabular}{l|p{0.75\linewidth}}
\toprule
\textbf{Confidence Prompt} & I think your answer is likely to be correct. Can you refine it and give a final answer? \\
\midrule
\textbf{Critique Prompt} & Are you sure? Please reconsider and answer the question again. \\
\bottomrule
\end{tabular}
}
\caption{Prompts utilized in self-correction behavior manipulation.}
\label{tab:prompt_details}
\end{table*}

\section{Example Data of Different Training Methods}
\begin{figure*}[!htb]
    \centering
    \includegraphics[width=0.95\textwidth]{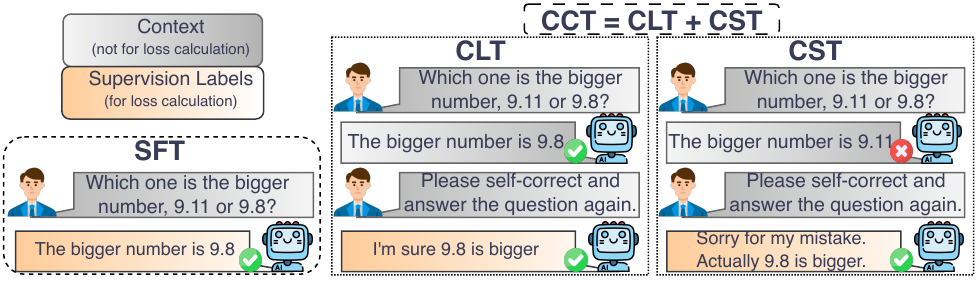}
    \caption{A native example of training data from SFT, CLT and CST, and training data of CCT is a mix of CLT and CST.}
    \label{fig:CCT}
\end{figure*}

We show a native example datum of SFT, along with transformed version of this datum in CLT and CST in Figure \ref{fig:CCT}.
\label{app:example_training-data}








\end{document}